\documentclass{article}

\PassOptionsToPackage{numbers, compress}{natbib}
\usepackage[final]{neurips_2020}

\usepackage[utf8]{inputenc} % allow utf-8 input
\usepackage[T1]{fontenc}    % use 8-bit T1 fonts
\usepackage{hyperref}       % hyperlinks
\usepackage{url}            % simple URL typesetting
\usepackage{booktabs}       % professional-quality tables
\usepackage{amsfonts}       % blackboard math symbols
\usepackage{nicefrac}       % compact symbols for 1/2, etc.
\usepackage{microtype}      % microtypography
\usepackage{amsthm}         % theorem
\usepackage{amssymb}        
\usepackage{amsfonts}
\usepackage{amsmath}        %math
\usepackage{soul}
\usepackage{bm}
\usepackage{xfrac}

\usepackage[table]{xcolor}
\usepackage{graphicx}
\usepackage{booktabs}
\usepackage{caption}
\usepackage{csquotes}
\usepackage{dsfont}

\newcommand{\cvec}[1]{\bm{#1}}

\newcommand{\cut}[1]{\text{cut}\hspace{-1.5px}\left(#1\right)}
\newcommand{\vol}[1]{\text{vol}\hspace{-1.5px}\left(#1\right)}

\renewcommand{\emptyset}{\varnothing}

\theoremstyle{definition}
\newtheorem{theorem}{Theorem}
\newtheorem{corollary}{Corollary}

\newcommand{\E}[1]{\mathbb{E}\left[#1\right]}

\newcommand{\gnn}{g}
 \newcommand{\Prob}[1]{P\left(#1\right)}
\newcommand{\clamp}[3]{\text{clamp}{\left(#1, #2, #3\right)}}

\title{Erd\H{o}s Goes Neural: an Unsupervised Learning Framework for Combinatorial Optimization on Graphs}

\author{%
  Nikolaos Karalias\\  
  EPFL\\
  \texttt{nikolaos.karalias@epfl.ch} \\ \And Andreas Loukas  \\
  EPFL\\
  \texttt{andreas.loukas@epfl.ch} \\
}

\begin{document}
%%%%%%%%%%%%%%%%V4 Changes:%%%%%%%%%%%%%%%%%%%%%%
% Rescale probabilistic penalty condition loss < epsilon *(1-t)
% Fix commas between citations in background section
%Constraints* instead of constrains on prob penalty section
%Period on the final sentence of supplement
%Max clique proof on supplement needs to say "weight" instead of "volume" and use indicator function instead of brackets.
%Add anonymous reviewers and Jegelka to acknowledgements.
%Add reference to the maxcut didactic example
%%%%%%%%%%%%%%%%%%%%%%%%%%%%%%%%%%%%%%%%%%%%%
\maketitle

\begin{abstract}
Combinatorial optimization (CO) problems are notoriously challenging for neural networks, especially in the absence of labeled instances. This work proposes an unsupervised learning framework for CO problems on graphs that can provide integral solutions of certified quality. 
Inspired by Erd\H{o}s' probabilistic method, we use a neural network to parametrize a probability distribution over sets. Crucially, we show that when the network is optimized w.r.t. a suitably chosen loss, the learned distribution contains, with controlled probability, a low-cost integral solution that obeys the constraints of the combinatorial problem. 
The probabilistic proof of existence is then derandomized to
decode the desired solutions. We demonstrate the efficacy of this approach to obtain valid  
solutions to the maximum clique problem and to perform local graph clustering. Our method achieves competitive results on both real datasets and synthetic hard instances. %Furthermore, it is shown that such a solution will obey the constraints of the combinatorial problem.  %a deterministic algorithm that approximate
% . However, providing guarantees on the quality of the solution after such a discretization process can be a challenging task. Recent approaches attempted to circumvent this difficulty by casting the unsupervised problem as a supervised one (by assuming a solution oracle) and/or by treating it as a sequential decision problem (e.g., based on reinforcement learning)---such solutions yield good results in certain situations, but are not always appropriate.    
% We assume a different approach: inspired by Erdos' probabilistic method, rather than asking the neural network to look for a fractional solution, we optimize the conditions for the existence of a good integral solution. Specifically, we construct a parametric distribution over sets and use the GNN to optimize the probability that at least one of those sets has a small cost. We then retrieve a set with the identified cost by means of a deterministic derandomization module.
\end{abstract}

% ============================================================
\section{Introduction}
% ============================================================
% \nikos{this intro is too huge, most people finish the intro before the halfway point in page 2}\andreas{It's going to be ok after we address the comments.}

Combinatorial optimization (CO) includes a wide range of computationally hard problems that are omnipresent in scientific and engineering fields. %, from biology and physics, to operations research and artificial intelligence (\nikos{citations}). 
Among the viable strategies to solve such problems are neural networks, which were proposed as a potential solution by \citet{hopfield1985neural}. % for the Travelling Salesman Problem (TSP). 
Neural approaches aspire to circumvent the worst-case complexity of NP-hard problems by only focusing on instances that appear in the data distribution.

%pff neeed to arrange this a bit better, need to come up with a transition
Since Hopfield and Tank, the advent of  deep learning has brought new powerful learning models, reviving interest in neural approaches for combinatorial optimization. A prominent example is that of graph neural networks (GNNs) \cite{gori2005new,scarselli2008graph}, whose success has motivated researchers to work on CO problems that involve graphs~\citep{joshi2019efficient, yolcu2019learning, khalil2016learning,  gasse2019exact,lemos2019graph,nowak2017note, bai2020fast, prates2019learning} or that can otherwise benefit from utilizing a graph structure in the problem formulation \citep{toenshoff2019run} or the solution strategy \citep{gasse2019exact}. The expressive power of graph neural networks has been the subject of extensive research~\cite{xu2019can,loukas2020hard, chen2020can,sato2019approximation, sato2020survey, barcelo2019logical, garg2020generalization}. 
Encouragingly, GNNs can be Turing universal in the limit~\cite{Loukas2020What}, which motivates their use as general-purpose solvers. %Many of the flagship CO problems are defined on graphs or can solved on graphs via polynomial time reductions. 

% \subsection{Key paradigms and recurring challenges}

Yet, despite recent progress, %the promising theoretical and experimental results, 
CO problems still pose a significant challenge to neural networks. %, as some of the longstanding obstacles remain unaddressed.
Successful models often rely on supervision, either in the form of labeled instances \cite{li2018combinatorial,selsam2018learning, joshi2019efficient} or of expert demonstrations \cite{gasse2019exact}. This success comes with drawbacks: obtaining labels for hard problem instances can be computationally infeasible~\citep{yehuda2020s}, and direct supervision can lead to poor generalization~\cite{joshi2019learning}. Reinforcement learning (RL) approaches have also been used for both classical CO problems~\cite{chen2019learning, yolcu2019learning, yao2019experimental, kool2018attention, deudon2018learning, khalil2017learning, bai2020fast} as well as for games with large discrete action spaces, like Starcraft \cite{vinyals2019grandmaster} and Go \cite{silver2017mastering}.
However, not being fully-differentiable, %while being able to mitigate some of the drawbacks of supervised methods by not requiring labels, 
they tend to be harder and more time consuming to train. 

%some extra stuff here
%To improve convergence, RL methods for CO problems have incorporated baselines \cite{kool2018attention} and curriculum learning \cite{yolcu2019learning}.

An alternative to these strategies is unsupervised learning, %In contrast to reinforcement learning where an agent learns a policy that aims to maximize its rewards in an environment, 
where the goal is to model the problem with a differentiable loss function whose minima represent the discrete solution to the combinatorial problem~\citep{smith1999neural,bianchi2019mincut, amizadeh2018learning, amizadeh2019pdp, toenshoff2019run, yao2019experimental}. 
Unsupervised learning is expected to aid in generalization, as it allows the use of large unlabeled datasets, and it is often envisioned to be the long term goal of artificial intelligence.  However, in the absence of labels, deep learning faces practical and conceptual obstacles.  
Continuous relaxations of objective functions from discrete problems are often faced with degenerate solutions or may simply be harder to optimize. Thus, successful training hinges on empirically-identified correction terms and auxiliary losses~\citep{bianchi2019mincut,amizadeh2019pdp, van1989improving}. Furthermore, it is especially challenging to decode valid (with respect to constraints) discrete solutions from the soft assignments of a neural network~\cite{li2018combinatorial, toenshoff2019run}, especially in the absence of complete labeled solutions \cite{selsam2018learning}.  
  
% \subsection{Our contributions}

\begin{figure}[t!]
    \centering
    \includegraphics[scale=0.6]{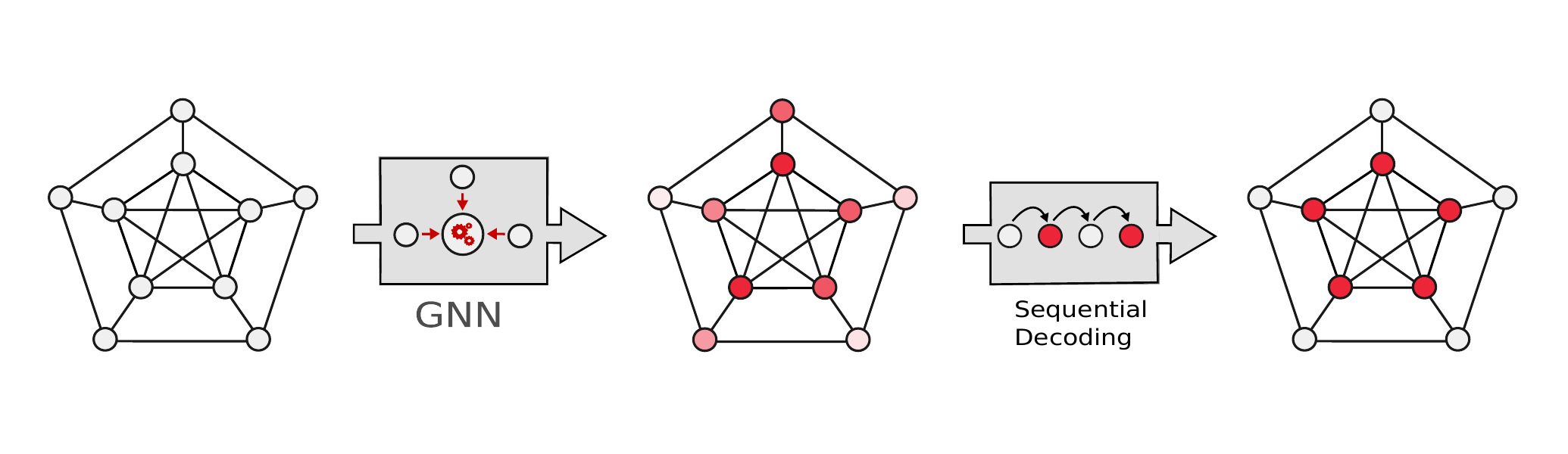}
    \caption{Illustration of the ``Erd\H{o}s goes neural''  pipeline. First, a differentiable loss is derived for a given problem using the probabilistic method. Next, a GNN is trained in an unsupervised way using the derived loss to output a probability distribution over the nodes, essentially providing a probabilistic certificate for the existence of a low cost feasible solution. At inference time, a discrete solution satisfying the certificate is obtained in a sequential and deterministic manner by the method of conditional expectation.\vspace{-5mm}}
    \label{fig:my_label}
\end{figure}

Our framework aims to overcome some of the aforementioned obstacles of unsupervised learning: \textit{it provides a principled way to construct a differentiable loss function whose minima are guaranteed to be low-cost valid solutions of the problem}. 
Our approach is inspired by Erd\H{o}s' probabilistic method and entails two steps: 
First, we train a GNN to produce a distribution over subsets of nodes of an input graph by minimizing a probabilistic penalty loss function. Successfully optimizing our loss is guaranteed to yield good integral solutions that obey the problem constraints. 
After the network has been trained, we employ a well-known technique from randomized algorithms to sequentially and deterministically decode a valid solution from the learned distribution. The procedure is schematically illustrated in Figure~\ref{fig:my_label}.

We demonstrate the utility of our method in two NP-hard graph-theoretic problems:  the \textit{maximum clique} problem~\cite{bomze1999maximum} and a \textit{constrained min-cut} problem~\cite{bruglieri2004cardinality, svitkina2011submodular} that can perform local graph clustering~\citep{andersen2006local,wang2017capacity}. 
In both cases, our method achieves competitive results against neural baselines, discrete algorithms, and mathematical programming solvers. Our method outperforms the CBC solver (provided with Google's OR-Tools), while also remaining competitive with the SotA commercial solver Gurobi 9.0~\citep{gurobi} on larger instances. Finally, our method outperforms both neural baselines and well-known local graph clustering algorithms in its ability to find sets of good conductance, while maintaining computational efficiency. \footnote{Code available at: https://github.com/Stalence/erdos\_neu}

\section{Related work and background}

% Historically, our general framework exhibits connections to the early works on neural networks for combinatorial optimization~\citep{hopfield1985neural, raghavan1988probabilistic, anderson1988neural, den1989graph}.  In fact, our probabilistic penalty formulation for the maximum clique problem from Section~\ref{subsec:probabilistic_loss}, leads to a loss function that is qualitatively similar to the energy functional presented in~\cite{ramanujam1995mapping}. These earlier works drew inspiration from statistical mechanics and formulated each problem in terms of energy minimization.

% Neural approaches to CO can be categorized based on whether they require access to labeled examples.

\subsection{Neural networks for combinatorial optimization}

Most neural approaches to CO are supervised. One of the first modern neural networks were the Pointer Networks \citep{vinyals2015pointer}, which utilized a sequence-to-sequence model for the travelling salesman problem (TSP).
Since then, numerous works have combined GNNs with various heuristics and search procedures to solve classical CO problems, such as quadratic assignment~\cite{nowak2017note}, graph matching~\citep{bai2018graph}, graph coloring~\citep{lemos2019graph}, TSP~\citep{li2018combinatorial,joshi2019efficient}, and even sudoku puzzles~\citep{palm2018recurrent}.
% 
% A notable work is the model in \citep{li2018combinatorial}, which uses a GNN within a maximum independent set heuristic and tree search to achieve strong performance on multiple CO problems, while also showing the ability to generalize. In \citep{palm2018recurrent}  a GNN is trained to be able to solve sudoku puzzles, while in \cite{joshi2019efficient} a GNN combined with beam search is used to solve TSP. Other applications of GNNs to CO problems include \cite{nowak2017note} for quadratic assignment problems, \cite{bai2018graph} for graph matching, \cite{lemos2019graph} for graph coloring, and \cite{prates2019learning} for the decision version of TSP.
% 
% \andreas{Out of place.} 
% \nikos{this was a concrete example of a loss function issue, it's for the unsupervised section but you told me that I was comparing too aggressively with others so i'm not sure what to do with it now}\andreas{Check if you can append the reference in one of the three categories in the end of the unsupervised challenges paragraph in the intro. } \nikos{I tried, take a look}
% 
Another fruitful direction has been the fusion with solvers. For example, Neurocore \citep{selsam2019guiding} incorporates an MLP to a SAT solver to enhance variable branching decisions, whereas \citet{gasse2019exact} learn branching approximations by a GNN and imitation learning. % to augment a B\&B solver.
Further, \citet{wang2019satnet} include an approximate SDP satisfiability solver as a neural network layer and \citet{vlastelica2019differentiation} incorporate exact solvers within a differentiable architecture by smoothly interpolating the solver's piece-wise constant output.
% 
% While the aforementioned works aim to add neural network functionality to solvers, the reverse direction %where a solver is integrated in an end to end differentiable pipeline 
% has also been explored. An example of that is \cite{wang2019satnet}, where an approximate SDP satisfiability solver is included as a neural network layer. Similarly, \citet{vlastelica2019differentiation} incorporate exact solvers within a differentiable architecture by smoothly interpolating the solver's piece-wise constant output.
% 
Unfortunately, the success of supervised approaches hinges on building large training sets with already solved hard instances, resulting in a chicken and egg situation. Moreover, %as shown by \citet{yehuda2020s}, %under mild complexity assumptions, 
since it is hard to efficiently sample unbiased and representative labeled instances of an NP-hard problem~\citep{yehuda2020s}, % 
% The latter implies that 
labeled instance generation is likely not a viable long-term strategy either.

% \nikos{not sure what to do with this paragraph}
% Solving NP-hard instances in order to obtain labels is not considered a viable long term strategy. An alternative that has been considered is to craft algorithms that generate labeled training instances. Alas, it was shown in \cite{yehuda2020s} that under mild computational complexity assumptions, it is impossible to have an efficient (polynomial-time) sampler that generates unbiased and representative labeled instances of an NP-hard problem. 

% \subsection{Solving CO problems without supervision}

Training neural networks without labels is generally considered to be more challenging. 
One possibility is to use RL: \citet{khalil2017learning} combine Q-Learning with a greedy algorithm and structure2vec embeddings to solve max-cut, minimum vertex cover, and TSP. Q-Learning is also used in \cite{bai2020fast} for the maximum common subgraph problem. On the subject of TSP, the problem was also solved with policy gradient learning combined with attention~\cite{kool2018attention, deudon2018learning, bello2016neural}. Attention is ubiquitous in problems that deal with sequential data, which is why it has been widely used with RL for the problem of vehicle routing \cite{gao2020learn, nazari2018reinforcement, peng2019deep, james2019online}. Another interesting application of RL is the work of \citet{yolcu2019learning}, where the REINFORCE algorithm is employed in order to learn local search heuristics for the SAT problem. This is combined with curriculum learning to improve stability during training. Finally, \citet{chen2019learning} use actor-critic learning to iteratively improve complete solutions to combinatorial problems. Though a promising research direction, deep RL methods are far from ideal, as they can be sample inefficient and notoriously unstable to train---possibly due to poor gradient estimates, dependence on initial conditions, correlations present in the sequence of observations, bad rewards, sub-optimal hyperparameters, or poor exploration~\citep{thrun1993issues,nikishin2018improving,rlblogpost,mnih2015human}. %They also require crafting an appropriate reward.

%perhaps due to the presence of noise in gradient estimators.
% \andreas{Any evidence against RL?}

The works that are more similar to ours 
% There has also been an increasing interest in unsupervised learning.  The works that are more similar to our approach 
are those that aim to train neural networks in a differentiable and end-to-end manner:    
\citet{toenshoff2019run} model CO problems in terms of a constraint language and utilize a recurrent GNN, where all variables  that coexist in a constraint can exchange messages. Their model is completely unsupervised and is suitable for problems that can be modeled as maximum constraint satisfaction problems. For other types of problems, like independent set, the model relies on empirically selected loss functions to solve the task. \citet{amizadeh2018learning, amizadeh2019pdp} train a GNN in an unsupervised manner to solve the circuit-SAT and SAT problems by minimizing an appropriate energy function. %The model for circuit SAT is geared towards problems with a DAG structure, while the PDP model improves upon that paradigm to accommodate general constraint satisfaction problems. 
Finally, \citet{yao2019experimental} train a GNN for the max-cut problem on regular graphs without supervision by optimizing a smooth relaxation of the cut objective and policy gradient.
 
Our approach innovates from previous works in the following ways: it enables training a neural network in an unsupervised, differentiable, and end-to-end manner, while also ensuring that identified solutions will be integral and will satisfy problem constraints. Crucially, this is achieved in a simple and mathematically-principled way, without resorting to continuous relaxations, regularization, or heuristic corrections of improper solutions. In addition, our approach does not necessitate polynomial-time reductions, but solves each problem directly. 

\subsection{Background: the probabilistic method}\label{subsec:prob_method}
The probabilistic method is a nonconstructive proof method pioneered by Paul Erd\H{o}s. It is used to demonstrate the existence of objects with desired combinatorial properties \cite{alon2004probabilistic,erdos1959graph, szegedy2013lovasz} but has also served as the foundation for important algorithms in the fields of computer science and combinatorial optimization \cite{moser2010constructive, raghavan1988probabilistic}.

Let us consider the common didactic example of the maximum cut problem on a simple undirected graph \cite{mitzenmacher2017probability}. The goal is to bipartition the nodes of the graph in such a way that the number of edges with endpoints in both partitions (i.e., the cardinality of the cut-set) is maximized. For simplicity we will refer to the cardinality of the cut-set as the cut.
Suppose we decide the bipartition based on a fair coin flip, i.e., we split the nodes of the graph by assigning them to a heads or a tails set. An edge belongs to the cut-set when its endpoints belong to different sets. This happens with probability $\sfrac{1\hspace{-1.2px}}{2}$, which implies that \textit{the expected cut} will be equal to half of the edges of the graph. Thus, by Markov's inequality and given that the cut is non-negative, it follows that there exists a bipartitioning that contains \textit{at least half} of the edges of the graph.

 To obtain such a solution deterministically, we will   utilize the method of conditional expectation \cite{raghavan1988probabilistic}: we sequentially visit every node $v_i$ in the graph and we compute the expected cut conditioned on $v_i$ belonging to the heads or tails set (together with all the decisions made until the $i$-th step) 
%  \begin{itemize}
%      \item $v_i$ belonging to the heads set; %, together with all the decisions made until the $i$-th step
%      \item $v_i$ belonging to the tails set. %, together with all all the decisions made until the $i$-th step
%  \end{itemize}
and add $v_i$ to the set (heads or tails) that yields smaller conditional expected cut.
Since the (conditional) expectation can only improve at every step, the sets recovered %is \textit{guaranteed to obtain at least the expected cut}. Therefore, we will have recovered a bipartitioning with 
are guaranteed to cut at least half the edges of the graph, as proved earlier.
 
 Our goal is to re-purpose this classic approach to tackle combinatorial optimization problems with deep learning.  Instead of using a naive probability assignment like in the maxcut example,  the probability distribution is learned by a GNN which allows us to obtain higher quality solutions. Additionally, we show how this argument may be extended to incorporate constraints within the learning paradigm.

\section{The Erd\H{o}s probabilistic method for deep learning}

We focus on combinatorial problems on weighted graphs $G = (V,E,w)$ that are modelled as constrained optimization problems admitting solutions that are node sets: %A constrained optimization problem with a feasible set of solutions has the following general form
\begin{align}
   \min_{S \subseteq V} \ f(S;G) \quad \text{subject to} \quad  S \in \Omega.
   \label{eq:combinatorial_problem}
\end{align}
Above, $\Omega$ is a family of sets having a desired property, such as forming a clique or covering all nodes. This yields a quite general formulation that can encompass numerous classical graph-theoretic problems, such as the maximum clique and minimum vertex cover problems. 
% 
% Above, $\Omega$ is a constraint set that imposes a desired structure on feasible solutions. % 
% Without significant loss of generality, in the following we focus in the case where $S$ contains nodes. %For instance, in the maximum clique problem feasible solutions may contain all nodes that form a clique. %, whereas in the maximum matching problem the optimization involves finding a set of edges that are not pairwise adjacent. 

% \subsection{In a nutshell}
\subsection{The \enquote{Erd\H{o}s Goes Neural} pipeline}

Rather than attempting to optimize the non-differentiable problem~\eqref{eq:combinatorial_problem} directly, we propose to train a GNN to identify distributions of solutions with provably advantageous properties. Our approach is inspired by Erd\H{o}s' probabilistic method, a well known technique in the field of combinatorics that is used to prove the existence of an object with a desired combinatorial property.

As visualized in Figure~\ref{fig:my_label}, our method consists of three steps:
\vspace{-2mm}
\begin{enumerate}
    \setlength\itemsep{.2em}
    \item Construct a GNN $\gnn_{\theta}$ that outputs a distribution $\mathcal{D} = \gnn_{\theta}(G)$ over sets.
    \item Train $\gnn_{\theta}$ to optimize the probability that there exists a valid $S^* \sim \mathcal{D}$ of small cost $f(S^*;G)$.
    \item Deterministically recover $S^*$ from $\mathcal{D}$ by the method of conditional expectation.
\end{enumerate}

There are several possibilities in instantiating $\mathcal{D}$. We opt for the simplest and suppose that the decision of whether $v_i \in S$ is determined by a Bernoulli random variable $x_{i}$ of probability $p_{i}$. The network can trivially parametrize $\mathcal{D}$ by computing $p_i$ for every node $v_i$. Keeping the distribution simple will aid us later on to tractably control relevant probability estimates. 

Next, we discuss how $\gnn_{\theta}$ can be trained (Section~\ref{subsec:probabilistic_loss}) and how to recover $S^*$ from $\mathcal{D}$ (Section~\ref{subsec:derand}).  
\subsection{Deriving a probabilistic loss function}
\label{subsec:probabilistic_loss}
% ============================================================

The main challenge of our method lies in determining how to tractably and differentiably train $\gnn_{\theta}$. Recall that our goal is to identify a distribution that contains low-cost and valid solutions. 

% ============================================================
\subsubsection{The probabilistic loss} 
\label{subsubsec:unconstrained}
% ============================================================

Aiming to build intuition, let us first consider the unconstrained case. To train the network, we construct a loss function $\ell(\mathcal{D}; G)$ that abides to the following property:
\begin{align}
    P( f(S;G) < \ell(\mathcal{D}; G)) > t \quad \text{with} \quad \mathcal{D} = \gnn_{\theta}(G).
\end{align}
Any number of tail inequalities can be used to instantiate such a loss, depending on the structure of $f$. If we only assume that $f$ is non-negative, Markov's inequality yields % 
$$
    \ell(\mathcal{D}; G) \triangleq \frac{\E{ f(S;G)}}{1-t} \quad \text{for any} \quad t\in [0,1). 
$$
If the expectation cannot be computed in closed-form, then any upper bound also suffices. 
% \andreas{Do we need that the function is positive and non-constant?}
% \nikos{I think it's necessary if we want a strict inequality from Markov. If you keep the equals sign (which messes up the probabilistic method) then you can drop that assumption}\andreas{I just checked and i think we only need non-negativity: Markov's gives : $P(X < a \E{X}) > 1 - 1/a$ for $a>0$. The equation above is obtained for $t = 1 - 1/a \in [0,1)$. } \nikos{i'll check this again}

The main benefit of approaching the problem in this manner is that the surrogate (and possibly differentiable) loss function $\ell(\mathcal{D}; G)$ can act as a certificate for the existence of a good set in the support of $\mathcal{D}$. To illustrate this, suppose that one has trained $\gnn_{\theta}$ until the loss is sufficiently small, say $\ell(\mathcal{D}; G) = \epsilon$. Then, by the probabilistic method, there exists with strictly positive probability a set $S^{*}$ in the support of $\mathcal{D}$ whose cost $f(S^{*};G)$ is at most $\epsilon$. %Therefore, as long as $\ell(\mathcal{D}; G)$ is efficiently computable, one may optimize $\gnn_\theta$ in a differentiable manner while guaranteeing that the loss corresponds to the cost of an integral solution.

% 
% which corresponds to the constrained optimization problem. Here, $\Omega$ denotes the feasible region for the constraint function $f_{c}$. on $S^{*}$ can be obtained via the method of conditional probabilities.

% =====================================================
\subsubsection{The probabilistic penalty loss} 
\label{subsubsec:constrained}
% =====================================================

To incorporate constraints, we take  inspiration from penalty methods in constrained optimization and add a term to the loss function that penalizes deviations from the constraint. 

% Our goal is to extend the framework described in the previous section to retrieve small-cost sets that also satisfy the constraints imposed by $\Omega$. Taking inspiration by penalty methods in constrained optimization, we add a term to the loss function that penalizes deviations from the constraint. 

Specifically, we define the probabilistic penalty function
$
    f_{p}(S;G) \triangleq f(S;G) + \cvec{1}_{S \notin \Omega} \, \beta,  \label{eq:probpen}
$
where $\beta$ is a scalar. % > \max_S \{f(S;G)\}$. 
The expectation of $f_p$ yields the probabilistic penalty loss: 
\begin{align}
     \ell(\mathcal{D}, G) \triangleq \E{f(S;G)} + P(S \notin \Omega) \, \beta.
\end{align}
We prove the following:

% The goal is to be able to use the probabilistic framework described in the previous section to retrieve small cost sets that also satisfy the constraints implied by $\Omega$. To achieve this, I utilize a probabilistic penalty method.
% \subsubsection{Probabilistic Penalty}
% Insert quick intro about penalty methods in optimization and cite something?

% \begin{theorem}
% Fix any $\beta > \max_S f(S;G)$. With strictly positive probability, there exists a set $S^{*} \sim \mathcal{D}$, for which 
% $
%     f(S^{*};G) < \ell(\mathcal{D}; G) \  \text{and} \  S^{*} \in \Omega,
% $
% under the condition that $f$ is non-negative.
% \label{theorem:probabilistic_penalty_certificate}
% \end{theorem}

\begin{theorem}
Fix any $\beta > \max_S f(S;G)$ and let $\ell(\mathcal{D}, G) =  (1-t)\epsilon < \beta$. With probability at least $t$, set $S^{*} \sim \mathcal{D}$ satisfies 
$$
    f(S^{*};G) < \ell(\mathcal{D}; G)/(1-t) \ \   \text{and} \ \ S^{*} \in \Omega,
$$
under the condition that $f$ is non-negative.
\label{theorem:probabilistic_penalty_certificate}
\end{theorem}

Hence, similar to the unconstrained case, the penalized loss acts as a certificate for the existence of a low-cost set, but now the set is also guaranteed to abide to the constraints $\Omega$. The main requirement for incorporating constraints is to be able to differentiably compute an upper estimate of the probability $P(S \notin \Omega)$. A worked out example of how $P(S \notin \Omega)$ can be controlled is provided in Section~\ref{subsec:max_clique}. 

% Theorem 1 suggests a natural way to incorporate constraints in the objective function via the probability of the constraint being violated by the set. 

% ============================================================
\subsubsection{The special case of linear box constraints}
\label{subsubsec:box_constraint}
% =====================================================

An alternative construction can be utilized when problem~\eqref{eq:combinatorial_problem} takes the following form: 
\begin{align}
    \min_{S \subseteq V} \ f(S;G) \quad \text{subject to} \quad \sum_{v_i \in S} a_i \in [b_l, b_h],
\end{align}
with $a_i$, $b_l$, and $b_h$ being non-negative scalars. 

We tackle such instances with a two-step approach. 
Denote by $\mathcal{D}^{0}$ the distribution of sets predicted by the neural network and let $p_1^{0}, \ldots, p_n^{0}$ be the probabilities that parametrize it. 
We rescale these probabilities such that the constraint is satisfied in expectation: 
$$
     \sum_{v_i \in V} a_i p_i  = \frac{b_l + b_h}{2}, \quad \text{where} \quad p_i = \clamp{c \, p_i^{0}}{0}{1}   \quad \text{and} \quad  c \in \mathbb{R}.
$$
Though non-linear, the aforementioned feasible re-scaling can be carried out by a simple iterative scheme (detailed in Section~\ref{app:iterative_scheme}). If we then proceed as in Section~\ref{subsubsec:unconstrained} by utilizing a probabilistic loss function that guarantees the existence of a good unconstrained solution, we have: 
% 
% The following theorem provides conditions for the existence of a low-cost valid set:
%
\begin{theorem}
Let $\mathcal{D}$ be the distribution obtained after successful re-scaling of the probabilities. For any (unconstrained) probabilistic loss function that abides to $ P(f(S;G) < \ell(\mathcal{D}; G)) > t$, set $S^* \sim \mathcal{D}$ satisfies 
$
    f(S^*;G) < \ell(\mathcal{D}; G) \  \text{and} \ \sum_{v_i \in S^*} a_i  \in [b_l, b_h],
$
with probability at least  $t - 2 \exp{\left(- (b_h - b_l)^2 / \sum_{i} 2 a_i^2 \right)}$.
%given that $t > 2 \exp{(- (b_h - b_l)^2 / (2 \sum_{v_i \in V} a_i^2))}$. % and $w = $.
\label{theorem:linear-box-constraint}
\end{theorem}
% 
% For instance, if $\ell(\mathcal{D}; G) \defeq \E{f(S; G)} = \epsilon$, then Markov's inequality implies that there exists a valid $S^*\sim \mathcal{D}$ with $f(S^*;G) < \frac{\epsilon}{1-2 \exp({- (b_h - b_l)^2 / w })}$. 
Section~\ref{subsec:min-cut} presents a worked-out example of how Theorem~\ref{theorem:linear-box-constraint} can be applied. % to the constrained minimum-cut problem. 

% \nikos{Does it make sense for the volume module to have its own section? }
% In this section we will describe an alternative approach to incorporating constraints in the objective function. Assuming a constraint is linearly dependent on the probabilities of the distribution $\mathcal{D}_{\bp}$, we define
% an iterative map $g: [0,1]^{|V|} \rightarrow [0,1]^{|V|}$  where at iteration $N$ we have $\mathcal{D}_{\bp}'$ such that
% % 
% \begin{align}
%     \E{f_{c}(S;G)}-t = 0.
% \end{align}
% % 
% with $t \in \mathbb{R}$ being the target value for the constraint.
% Then, our objective becomes
% \begin{align}
%     \underset{\mathcal{D}_{\bp}'}{\text{minimize}} \quad \E{f_{o}(S;G)\vert  \E{f_{c}(S;G)} = t}.
% \end{align}

% The iteration at step $i$ is given by
% \begin{align}
%     \bp^{i} \leftarrow \min(a^{i} \bp^{i-1},1), \quad i=1,2 \dots N, 
% \end{align}
% with $a^{i} \in \mathbb{R}$ a scaling factor and $\min$ taken as elementwise minimum. At a given step i, we obtain a distribution $\mathcal{D}_{\bp}^{i}$.
% The scaling factor $a^{i}$ can be calculated as
% \begin{align}
%     a^{i} = \frac{t}{ \mathbf{E}_{\mathcal{D}_{\bp^{i}}} [{f_{c}(S;G)}]}.
% \end{align}

% Write about derandomization that accounts for the constraint

% ============================================================
\subsection{Retrieving integral solutions}
\label{subsec:derand}
% ===========================================================

A simple way to retrieve a low cost integral solution $S^*$ from the learned distribution $\mathcal{D}$ is by monte-carlo sampling. Then, if $S^*\sim \mathcal{D}$ with probability $t$, the set can be found within the first $k$ samples with probability at least $1 - (1-t)^k$.  However, our goal is to deterministically obtain $S^*$ so we will utilize the method of conditional expectation that was introduced in Section \ref{subsec:prob_method}.

Let us first consider the unconstrained case.
Given $\mathcal{D}$, the goal is to identify a set $S^*$ that satisfies $f(S^*;G) \leq \E{f(S; G)}$.
To achieve this, one starts by sorting $v_1, \ldots, v_n$ in  order of decreasing probabilities $p_i$.
Let $S_{\text{reject}}= \emptyset$ be the set of nodes not accepted in the solution.
Set $S^* = \emptyset$ is then iteratively updated one node at a time, with $v_i$ being included to $S^*$ in the $i$-th step if $\E{f(S ;G)\ |\ S^* \subset S, \ S \cap S_{\text{reject}} = \emptyset, \  \text{ and } v_{i} \in S} < \E{f(S;G)\ | \ S^* \subset S, \ S \cap S_{\text{reject}} = \emptyset, \ \text{ and } v_{i} \notin S}$.
% % 
% \begin{align}
%     S^* = \begin{cases} 
%     S^* & \text{if } \ \E{f(S ;G)\ |\ S^* \subset S  \text{ and } v_{i} \in S} > \E{f(S;G)\ | \ S^* \subset S \text{ and } v_{i} \notin S}, \\
%     \\
%     S^* \cup \{ v_{i} \} & \text{otherwise.} 
%     \end{cases}\notag 
% \end{align}
% 
This sequential decoding works because the conditional expectation never increases. %, after all nodes have been considered $S^{*}$ obeys $f(S^{*};G) \leq \E{f(S; G)}.$

% : Starting with the empty set $S=\{\emptyset \}$, it is an iterative process where we fix an ordering of the nodes  and then for each node $v_{i}$,  its inclusion in the set $S$ is decided by calculating the expectation of the objective function conditioned on the membership of $v_{i}$. Denote with $S^{(i)}$ the set at step i, then
% \begin{align}
%     S^{(i+1)} = \begin{cases} S^{(i)} & \text{if }  \E{f(S^{(i)} ;G)|v_{i} \in S^{(i)}} > \E{f(S^{(i)};G)|v_{i} \notin S^{(i)}} \\
%     \\
%     S^{(i)} \cup \{ v_{i} \} & \text{if }  \E{f(S^{(i)};G)|v_{i} \in S^{(i)}} \leq \E{f(S^{(i)};G)|v_{i} \notin S^{(i)}}
%     \end{cases}
% \end{align}
% % \begin{align}
% %  S^{(i+1)}= \underset{S}{\text{argmin}} \left(\E{ f_{}(S^{(i+1)};G)| v_{i} \in S}, \E{ f_{}(S^{(i+1)};G)| v_{i} \not\in S} \right), \; \forall i.
% % \end{align}
% Observe from the expression above that $\E{f(S^{(i+1)};G)}\leq \E{f(S^{(i)};G)}$.
% This guarantees that after $N=|V|$ steps when all nodes have been considered, the obtained set $S^{*}=S^{(N)}$ obeys $    f_{}(S^{*};G) < \epsilon. \label{eq:quality} $

In the case of the probabilistic penalty loss, the same procedure is applied w.r.t. the expectation of $f_p(S; G)$. The latter ensures that the decoded set will match the claims of Theorem~\ref{theorem:probabilistic_penalty_certificate}. % i.e., $S^* \in \Omega$ and $f(S^*; G) \leq \ell(\mathcal{D}; G) $. 
For the method of Section~\ref{subsubsec:box_constraint}, a sequential decoding can guarantee either that the cost of $f(S^*;G)$  is small or that the constraint is satisfied. %However, it is possible to manually control whether the constraint is obeyed with each node addition during the iteration process as we will demonstrate later on.

\section{Case studies} 
% ============================================================

This section demonstrates how our method can be applied to two well known NP-hard problems: the \textit{maximum clique}~\cite{bomze1999maximum} and the \textit{constrained minimum cut}~\cite{bruglieri2004cardinality} problems. %We chose two classic NP-hard problems that are known to be challenging and that can allow us to test the two different approaches to incorporating constraints.

\subsection{The maximum clique problem}
\label{subsec:max_clique}
% ============================================================

A clique is a set of nodes such that every two distinct nodes are adjacent. The maximum clique problem entails identifying the clique of a given graph with the largest possible number of nodes:
% 
% A set $S \subseteq V$ of nodes %be a set of nodes of volume $w(S) = \sum_{v_i, v_j \in S} w_{ij}$.
% 
% Let $G_S = (S, E_{S}, w_S)$ be the subgraph induced by set $S$ on a weighted graph $G=(V,E,w)$.
% 
% Set $S$ 
% corresponds to a clique if every node pair $v_i, v_j \in S$ is connected by an edge $(v_i,v_j) \in E$. 
% 
% A clique $C_{S}$ in an unweighted graph $G$ is a subset of nodes $S$ for which the  induced subgraph is complete. The clique number of a graph $G$, commonly denoted as $\omega(G)$, is the cardinality of the largest clique in $G$. 
% In the maximum clique problem one has to find the largest subset $S$ that induces a clique $G_S$. 
% 
% We formulate the maximum clique problem as follows: 
% 
\begin{align}
    \min_{S\subseteq V} - w(S) \quad \text{subject to} \quad S \in \Omega_\text{clique},  
    \label{eq:maxclique_2}
\end{align}
with $\Omega_\textit{clique}$ being the family of cliques of graph $G$ and 
$w(S) = \sum_{v_i, v_j \in S} w_{ij}$ 
being the weight of $S$. Optimizing $w(S)$ is a generalization of the standard cardinality formulation to weighted graphs. For simple graphs, both weight and cardinality formulations yield the same minimum. 
% 
% Term $\gamma$ is any upper bound on the max-clique size that renders the objective is non-negative. 

% \begin{align}
%     \max_{S\subseteq V} \, |S| \quad \text{subject to} \quad S \in \Omega_\text{clique},
%     \label{eq:maxclique}
% \end{align}
% % 
% with $\Omega_\textit{clique}$ being the family of sets $S$ whose induced subgraph is a clique.
% % 
% % % ============================================================
% % \subsubsection{Probabilistic Penalty Method} 
% % % ============================================================
% % 
% It will be convenient to consider the following equivalent formulation:
% % 
% \begin{align}
%     \min_{S\subseteq V} \, \gamma - w(G_S) \quad \text{subject to} \quad S \in \Omega_\text{clique},  
%     \label{eq:maxclique_2}
% \end{align}
% % 
% where now one minimizes the negative volume $w(G_{S}) = \sum_{e_{ij} \in E_{S}} w_{ij} x_{i}x_{j}$ of the clique. Apart from casting it as a minimization problem (as is common in deep learning), problem~\eqref{eq:maxclique_2} provides a convenient generalization to weighted graphs.
% % 
% On the other hand, $\gamma$ is any upper bound on the max-clique size that is introduced to ensure that the objective is non-negative. 
 
We can directly apply the ideas of Section~\ref{subsubsec:constrained} to derive a probabilistic penalty loss:

\begin{corollary}
Fix positive constants $\gamma$ and $\beta$ satisfying $\max_{S} w(S) \leq \gamma \leq \beta$ and let $w_{ij} \leq 1$. If 
\begin{align*}
    \ell_{\text{clique}}(\mathcal{D}, G) 
    &\triangleq \gamma - (\beta+1) \sum_{(v_i,v_j) \in E} w_{ij} p_{i} p_{j} +\frac{\beta}{2} \sum_{v_i\neq v_j } p_i p_j \leq \epsilon (1-t)
\end{align*}
then, with probability at least $t$, set $S^* \sim \mathcal{D}$ is a clique of weight $ w(S^*)>\gamma- \epsilon$. 
\label{corrolary:max_clique}
\end{corollary}
% 
% The proof can be found in the appendix.
% 
The loss function $\ell_{\text{clique}}$ can be evaluated in linear time w.r.t. the number of edges of $G$ by rewriting the rightmost term as $\sum_{v_i\neq v_j } p_i p_j = (\sum_{v_i \in V} p_i)^2 - \sum_{(v_i,v_j) \in E} 2 p_i p_j $.

A remark. One may be tempted to fix $\beta \to \infty$, such that the loss does not feature any hyper-parameters.
% 
% \begin{align*}
%     \lim_{\beta \to \infty} \ell_{\text{clique}}(\mathcal{D}, G) 
%       &\propto -\sum_{(v_i,v_j) \in E} w_{ij} p_{i}p_{j} + \frac{1}{2}\sum_{\substack{v_i \neq v_j}} p_{i}p_{j}.
% \end{align*}
% 
However, with mini-batch gradient descent it can be beneficial to tune the contribution of the two terms in the loss to improve the optimization. This was also confirmed in our experiments, where we selected the relative weighting according to a validation set. 

\paragraph{Decoding cliques.} 
After the network is trained, valid solutions can be decoded sequentially based on the procedure of Section~\ref{subsec:derand}. The computation can also be sped up by replacing conditional expectation evaluations (one for each node) by a suitable upper bound. Since the clique property is maintained at every point, %This can easily be enforced during run-time with a simple check. 
we can also efficiently decode cliques by sweeping nodes (in the order of larger to smaller probability) and only adding them to the set when the clique constraint is satisfied.

\subsection{Graph partitioning}
\label{subsec:min-cut}
% ============================================================

% A cut is a partition of the node set $V$ of a graph $G = (V,E,w)$ into complement subsets $S \text{ and } \bar{S}$. 
 
The simplest partitioning problem is the minimum cut: find set $S \subset V$ such that $\cut{S} = \sum_{v_i\in S, \ v_j\notin {S}} w_{ij}$ is minimized. 
Harder variants of partitioning aim to provide control on partition balance, as well as cut weight.
We consider the following constrained min-cut problem: 
\begin{align*}
    \underset{S}{\min} \ \cut{S} \quad \text{subject to} \quad \vol{S} \in [v_l, v_h], 
\end{align*}
where the volume $\vol{S}= \sum_{v_i \in S} d_{i}$ of a set is the sum of the degrees of its nodes. 
% 
% The volume constrained min-cut is a simple extension of the cardinality constrained min-cut, which is known to be NP-hard \cite{bruglieri2004cardinality}.

% \nikos{maybe that sentence needs to go?}
% Constrained minimization of submodular functions like the cut is generally NP-hard, even for simple lower bound constraints on the cardinality \cite{iyer2013fast}.

The above can be shown to be NP-hard \cite{iyer2013fast} and exhibits strong connections with other classical formulations: it is a volume-balanced graph partitioning problem~\citep{andreev2006balanced} and can be used to minimize graph conductance~\citep{chung1997spectral} by scanning through solutions in different volume intervals and selecting the one whose cut-over-volume ratio is the smallest (this is how we test it in Section~\ref{sec:empirical}).

% The weight of a cut is denoted by $\cut{S} = \sum_{v_i\in S \text{ and } v_j\notin {S}} w_{ij}$ and the volume of $S$ is $\text{vol}(S)= \sum_{i \in S} d_{i}$, where $d_{i}$ is the degree of $v_{i}$. 

% Denote the cut edge set by $E_S  \defeq \{ (v_i,v_j) \in E|  v_i\in S \text{ and } v_j\notin {S}\}$ and its weight by $\cut{S} = \sum_{{(i,j)} \in E} w_{ij}$. The volume of $S$ is defined as $\text{vol}(S)= \sum_{i \in S}d_{i}$, where $d_{i}$ is the degree of vertex $v_{i}$.

% In the minimum cut problem one has to find a set $S \subset V$, such that  $w(B_{S})$  is minimized.  In the volume constrained variant of the problem, one has to solve the minimum cut problem under the constraint that $vol(S)=t$, for some target volume $t$.

% Since the volume constraint is a linear box constraint,  
We employ the method described in Section~\ref{subsubsec:box_constraint} to derive a probabilistic loss function:  
\begin{corollary}
Let the probabilities $p_1, \ldots, p_n$ giving rise to $\mathcal{D}$ be re-scaled such that 
$\sum_{v_i \in V} d_i p_i =  \frac{v_l + v_h}{2}$ and, further, fix 
$
    \ell_{\text{cut}}(\mathcal{D}; G) \triangleq  \sum_{v_i \in V} d_i p_i - 2\sum_{(v_i,v_j) \in E} p_{i}p_{j} w_{ij}.
$
Set $S^* \sim \mathcal{D}$ satisfies 
\begin{align*}
    \cut{S^*} < {\ell_{\text{cut}} (\mathcal{D}; G)}/{(1 - t)} \quad \text{and} \quad \vol{S^*} \in [v_l, v_h],
\end{align*}
with probability at least $t - 2 \exp{\left(- (v_h - v_l)^2 / \sum_{i} 2 d_i^2 \right)}$.
\label{cor:min_cut}
\end{corollary}
The derived loss function $\ell_{\text{cut}}$ can be computed efficiently on a sparse graph, as its computational complexity is linear on the number of edges.

% The constrained version of the optimization problem can be written as
% \begin{align*}
% \underset{S}{\min} \ w(B_{S}), \ S \in \Omega. 
% \end{align*}
% The family of solutions $\Omega$ is such that for every $S \in \Omega, \text{vol}(S) = t$. The constraint is incorporated in the objective by minimizing the expectation of the objective function, conditioned on the contraint being satisfied on expectation, which yields the following loss
% \begin{align}
%     \ell_{cut} = \E{cut({S}))|\E{\text{vol}(S)}=t} = \sum_{(i,j) \in E}w_{ij}(p_{i}+p_{j}) - 2 \sum_{(i,j) \in E}w_{ij}p_{i}p_{j}.
% \end{align}

% This function is non-negative and therefore the probablistic method can be used to show that if $\ell_{cut}<\epsilon$ then there exists a set $S$ with $w(B_{S})<\epsilon$. Conditioning on the expected volume happens through an iterative rescaling of the probabilities. 

\paragraph{Decoding clusters.} 
Retrieving a set that respects Corollary~\ref{cor:min_cut} can be done by sampling. Alternatively, the method described in Section~\ref{subsec:derand} can guarantee that the identified cut is at most as small as the one certified by the probabilistic loss. In the latter case, the linear box constraint can be practically enforced by terminating before the volume constraint gets violated. 

% \andreas{the following is out of place. }
% Here we should note that the solutions are conditioned on the input node being in the set. That is, if $ \cvec{p} = \text{GNN}(\boldsymbol{\delta}_{i};G)$ are the output probabilities, then  $v_{i} \in S$ . This helps enforce local clusters that are connected, which is motivated by the fact that locally biased clustering methods have been shown to produce better quality clusters compared to their global counterparts in real world applications \cite{jeub2015think}.

% ============================================================
\section{Empirical evaluation}
\label{sec:empirical}
% ============================================================
We evaluate our approach in its ability to find large cliques and partitions of good conductance. 

\subsection{Methods}

We refer to our network as Erd\H{o}s' GNN, paying tribute to the pioneer of the probabilistic method that it is inspired from. Its architecture comprises of multiple layers of the Graph Isomorphism Network (GIN) \cite{xu2018powerful} and a Graph Attention (GAT) \cite{velivckovic2017graph} layer. %We found that using a multi-head GAT layer was beneficial to the optimization of the network. 
Furthermore, each convolution layer was equipped with skip connections, batch normalization and graph size normalization~\citep{dwivedi2020benchmarking}. % which were shown to be beneficial for the training convergence of GNNs~\citep{dwivedi2020benchmarking}. 
In addition to a graph, we gave our network access to a one-hot encoding of a randomly selected node, which encourages locality of solutions, allows for a trade-off between performance and efficiency (by rerunning the network with different samples), and helps the network break symmetries~\citep{seo2019discriminative}. 
Our network was trained with mini-batch gradient descent, using the Adam optimizer~\citep{kingma2014adam} and was implemented on top of the pytorch geometric API~\citep{fey2019fast}.

\emph{Maximum clique.}  
We compared against three neural networks, three discrete algorithms, and two integer-programming solvers: 
The neural approaches comprised of \textit{RUN-CSP}, \textit{Bomze GNN}, and \textit{MS GNN}. The former is a SotA unsupervised network %for binary constraint satisfaction problems; it 
incorporating a reduction to independent set and a post-processing of invalid solutions with a greedy heuristic. The latter two, though identical in construction to Erd\H{o}s' GNN, were trained based on standard smooth relaxations of the maximum clique problem with a flat 0.5-threshold discretization~\citep{motzkin1965maxima, bomze1997evolution}. 
Since all these methods can produce multiple outputs for the same graph (by rerunning them with different random node attributes), we fix two time budgets for RUN-CSP and Erd\H{o}s' GNN, that we refer to as ``fast" and ``accurate" and rerun them until the budget is met (excluding reduction costs). On the other hand, the Bomze and MS GNNs are rerun 25 times, since further repetitions did not yield relevant improvements.
We considered the following algorithms: the standard \textit{Greedy MIS Heur.} which greedily constructs a maximal independent set on the complement graph, \textit{NX MIS approx.}~\citep{boppana1992approximating}, and \textit{Toenshoff-Greedy}~\citep{toenshoff2019run}.  
Finally, we formulated the maximum clique in integer form~\citep{bomze1999maximum} and solved it with \textit{CBC}~\cite{johnjforrest_2020_3700700} and \textit{Gurobi} 9.0~\citep{gurobi}, an open-source solver provided with Google's OR-Tools package and a SotA commercial solver. We should stress that our evaluation does not intend to establish SotA results (which would require a more exhaustive comparison), but aims to comparatively study the weaknesses and strengths of key unsupervised approaches. 
% 
% The solvers, we describe the problem using an integer programming formulation .
% For  the neural baselines, we implemented  MS GNN and  Bomze GNN, which are architecturally similar to Erd\H{o}s' GNN but optimize the two well known continuous formulations of the problem \cite{motzkin1965maxima, bomze1997evolution} and use a flat 0.5 threshold for discretization.
% For the hard maximum clique instances, we only compared to Toenshoff-Greedy, RUN-CSP, and Gurobi, as the rest of the baselines did not yield meaningful results.

\emph{Local partitioning.}
We compared against two neural networks and four discrete algorithms. To the extent of our knowledge, no neural approach for constrained partitioning exists in the literature.  Akin to maximum clique, we built the \textit{L1 GNN} and \textit{L2 GNN} to be identical to Erd\H{o}s' GNN and trained them based on standard smooth $\ell_1$ and $\ell_2$ relaxations of the cut combined with a volume penalty. 
On the other hand, a number of algorithms are known for finding small-volume sets of good conductance. We compare to well-known and advanced algorithms~\cite{fountoulakis2018short}: \textit{Pagerank-Nibble}~\cite{andersen2006local}, Capacity Releasing Diffusion (\textit{CRD})~\cite{wang2017capacity}, Max-flow Quotient-cut Improvement (\textit{MQI}) \cite{lang2004flow} and \textit{Simple-Local}~\cite{10.5555/3045390.3045595}. 

% ========================================
\subsection{Data}
% ========================================

Experiments for the maximum clique were conducted in the IMDB, COLLAB~\citep{KKMMN2016,yanardag2015deep} and TWITTER \citep{snapnets} datasets, listed in terms of increasing graph size. %, containing respectively ego graphs of actors in movies, scientific collaboration networks, and ego networks of twitter users.
Further experiments were done on graphs generated from the RB model~\citep{xu2007random}, that has been specifically designed to generate challenging problem instances. We worked with three RB datasets: a training set containing graphs of up to 500 nodes~\citep{toenshoff2019run}, a newly generated test set containing graphs of similar size, and a set of instances that are up to 3 times larger~\citep{xu2007benchmarks,li2018combinatorial,toenshoff2019run}.   
% 
% For the training set, we used the same training dataset that was generated in the work by \citet{toenshoff2019run} to train our models. It consists of hard instances of up to 500 nodes. We tested on 3 settings using the same models: the training set, a test set in the same distribution, a test set of large instances. For the test set, we generated 500 graphs within the same graph size range as the training set and compared on those. Furthermore, we tested on the \citet{xu2007benchmarks} large maximum-clique  instances. It is a set of hard instances that are up to 3 times larger than the training set and have been commonly used as benchmarks in the literature  \cite{li2018combinatorial, toenshoff2019run}.  \andreas{to here}
% 
On the other hand, to evaluate partitioning, we focused on the FACEBOOK~\cite{traud2012social}, TWITTER, and SF-295 \cite{yan2008mining} datasets, with the first being a known difficult benchmark. %FACEBOOK consists of social graphs, each one representing friend groups of a major American university. SF-295 is a collection of molecular graphs from cancer screenings of the central nervous system. 
% 
% \paragraph{Hard satisfiable instances (RB Model)}For the training set, we used the same training dataset that was generated in the work by \citet{toenshoff2019run} to train our model. It consists of hard instances of up to 500 nodes. For the test set, we generated additional graphs from the same distribution. Furthermore, we tested on the \citet{xu2007benchmarks} large maximum-clique  instances. It is a set of hard instances that are up to 3 times larger than the training set and have been commonly used as benchmarks in the literature  \cite{li2018combinatorial, toenshoff2019run}. 
% 
More details can be found in the Appendix. %~\ref{app:details}.

% \textit{Metrics.} For the maximum clique problem, we report the approximation ratio, %, a real number in the $[0,1]$ interval 
% i.e., the ratio of the solution's cost over the optimal cost. We also report the average evaluation time per graph. For the partitioning experiments, we report the standard measure of local conductance, defined as $\phi(S) = \cut{S}/\vol{S}$. 

\textit{Evaluation.} We used a 60-20-20 split between training, validation, and test for all datasets, except for the RB model data (details in paragraph above). Our baselines often require the reduction of maximum clique to independent set, which we have done when necessary. The reported time costs factor in the cost of reduction. During evaluation, for each graph, we sampled multiple inputs, %$\boldsymbol{\delta}_{i}$, 
obtained their solutions, and kept the best one. This was repeated for all neural approaches and local graph clustering algorithms. Solvers were run with multiple time budgets.

%evaluation details
% For the neural network baselines, we incorporated the constraint using a simple linear penalty function on the volume. For Pagerank-Nibble, we randomly select a lower bound for the volume that is at least 10 \% of the total graph volume.

\begin{table}
\centering
\resizebox{0.87\textwidth}{!}{
\begin{tabular}{l c c c}    \toprule
\emph{}        & {IMDB} & COLLAB & TWITTER  \\\midrule
 Erd\H{o}s' GNN (fast)      & 1.000 (0.08 s/g)  & 0.982 $\pm$  0.063 (0.10 s/g)        & \textbf{0.924 $\pm$ 0.133 (0.17 s/g)}
  \\ 
 Erd\H{o}s' GNN (accurate)      & 1.000 (0.10 s/g)     & 0.990 $\pm $  0.042  (0.15 s/g)         & 0.942 $\pm$  0.111 (0.42 s/g)  \\ 
RUN-CSP (fast)       & 0.823 $\pm$ 0.191 (0.11 s/g)    & 0.912 $\pm$ 0.188   (0.14 s/g)          & 0.909 $\pm$ 0.145 (0.21 s/g) \\
RUN-CSP (accurate)   & 0.957 $\pm$ 0.089  (0.12 s/g)   & 0.987 $\pm$ 0.074    (0.19 s/g)         & 0.987 $\pm$ 0.063 (0.39 s/g)  \\
Bomze GNN            & 0.996  $\pm$ 0.016 (0.02 s/g)   & \textit{0.984 $\pm$ 0.053  (0.03 s/g)}  & \textit{0.785 $\pm$ 0.163 (0.07 s/g)} \\
MS GNN               & 0.995 $\pm$ 0.068 (0.03 s/g)    & \textit{0.938 $\pm$ 0.171 (0.03 s/g)}   & \textit{0.805 $\pm$ 0.108 (0.07 s/g)}  \\
% RUN-CSP (32)         & 1.000                   & 0.996 $\pm$ 0.048           & 0.998 $\pm$ 0.007  \\
% RUN-CSP (8)          & 0.990 $\pm$ 0.051       & 0.993 $\pm$ 0.059           & 0.996 $\pm$ 0.018  \\ 
\midrule
NX MIS approx.       & 0.950 $\pm$ 0.071    (0.01 s/g)  & 0.946 $\pm$ 0.078   (1.22 s/g)        & 0.849 $\pm$ 0.097 (0.44 s/g) \\ 
Greedy MIS Heur.   & 0.878 $\pm$ 0.174 (1e-3 s/g)       & 0.771 $\pm$ 0.291 (0.04 s/g)          & 0.500 $\pm$ 0.258 (0.05 s/g) \\ 
Toenshoff-Greedy   & 0.987 $\pm$ 0.050 (1e-3 s/g)       & 0.969 $\pm$ 0.087 (0.06 s/g)        & \textbf{0.917 $\pm$ 0.126 (0.08 s/g)} \\ \midrule
CBC (1s)             & 0.985 $\pm$ 0.121  (0.03 s/g)    & 0.658 $\pm$ 0.474 (0.49 s/g)          & 0.107 $\pm$ 0.309 (1.48 s/g) \\
CBC (5s)             & 1.000 (0.03 s/g)                 & 0.841 $\pm$ 0.365  (1.11 s/g)        & 0.198 $\pm$ 0.399 (4.77 s/g) \\ 
% CBC (20s)            & 1.000                   & 0.984 $\pm$ 0.122           & 0.534 $\pm$ 0.482 \\
Gurobi 9.0 (0.1s)        & \textbf{1.000    (1e-3 s/g)}   & 0.982 $\pm$ 0.101    (0.05 s/g)       & 0.803 $\pm$ 0.258 (0.21 s/g) \\
Gurobi 9.0 (0.5s)        & 1.000    (1e-3 s/g)            & 0.997 $\pm$ 0.035   (0.06 s/g)        & 0.996 $\pm$ 0.019  (0.34 s/g)\\
Gurobi 9.0 (1s)          & 1.000    (1e-3 s/g)            & 0.999 $\pm$ 0.015   (0.06 s/g)        & \textbf{1.000 (0.34 s/g)} \\
Gurobi 9.0 (5s)          & 1.000     (1e-3 s/g)           & \textbf{1.000    (0.06 s/g)}           & 1.000 (0.35 s/g) \\
% Gurobi (20s)         & 1.000                   & 1.000                       & 1.000  \\
\bottomrule \\
\end{tabular} }
\caption{Test set approximation ratios for all methods on real-world datasets. For solvers, time budgets are listed next to the name. Pareto-optimal solutions are indicated in bold, whereas italics indicate constraint violation (we report the results only for correctly solved instances). \vspace{-.4cm}}
\label{table:max-clique-real}
\end{table}

% ============================================================
\subsection{Results: maximum clique}
% ============================================================

Table~\ref{table:max-clique-real} reports the test set approximation ratio, i.e., the ratio of each solution's cost over the optimal cost. %achieved by each method. 
For simple datasets, such as IMDB, most neural networks achieve similar performance and do not violate the problem constraints. On the other hand, the benefit of the probabilistic penalty method becomes clear on the more-challenging Twitter dataset, where training with smooth relaxation losses yields significantly worse results and constraint violation in at least 78\% of the instances (see Appendix). Erd\H{o}s' GNN always respected constraints. Our method was also competitive w.r.t. network RUN-CSP and the best solver, consistently giving better results when optimizing for speed (``fast"). %whereas Guwith both methods performing similarly on a bigger time-budget. 
The most accurate method overall was Gurobi, which impressively solved all instances perfectly given sufficient time. As observed, Gurobi has been heavily engineered to provide significant speed up w.r.t. CBC. Nevertheless, we should stress that both solvers scale poorly with the number of nodes and are not viable candidates for graphs with more than a few thousand nodes.

% Since all neural baselines can produce multiple outputs for the same graph (by rerunning them with different random node attributes), we fix two time budgets for RUN-CSP and Erd\H{o}s' GNN, that we refer to as ``fast" and ``accurate". We compute for each one the number of samples that matches the budget. For the continuous relaxation nets we compute 25 samples on all graphs, which albeit fast, violate the constraints up to 85 \% on twitter and up to 15\% on COLLAB. There was at most 1 \% violation on IMDB so the results remain valid on that dataset. Our model performs better than RUN-CSP on smaller time budgets but RUN-CSP shows better performance on larger out-of-distribution instances. Our Meth

Table~\ref{table:max-clique-hard} tests the best methods on hard instances. We only provide the results for Toenshoff-Greedy, RUN-CSP, and Gurobi, as the other baselines did not yield meaningful results.
Erd\H{o}s' GNN can be seen to be better than RUN-CSP in the training and test set and worse for larger, out of distribution, instances. However, both neural approaches fall behind the greedy algorithm and Gurobi, especially when optimizing for quality. The performance gap is pronounced for small instances but drops significantly for larger graphs, due to Gurobi's high computational complexity. It is also interesting to observe that the neural approaches do better on the training set than on the test set. Since both neural methods are completely unsupervised, the training set performance can be taken at face value (the methods never saw any labels). Nevertheless, the results also show that both methods partially overfit the training distribution. The main weakness of Erd\H{o}s' GNN is that its performance degrades when testing it in larger problem instances. Nevertheless, it is encouraging to observe that even on graphs of at most 1500 nodes, both our ``fast'' method and RUN-CSP surpass Gurobi when given the same time-budget. We hypothesize that this phenomenon will be more pronounced with larger graphs. %Our hypothesis is that improved generalization beyond the distribution could be possible, e.g., with more optimized architectures and curriculum learning, but we lack empirical evidence to make conclusive statements.       

\begin{table}[]
\centering
\resizebox{0.86\textwidth}{!}{%
\begin{tabular}{l c c c} \toprule
% \begin{tabular}{m{0.3\textwidth}m{0.3\textwidth}m{0.3\textwidth}m{0.3\textwidth}} \toprule
                            & Training set                           & Test set                              & Large Instances    \\ \midrule 
Erd\H{o}s' GNN (fast)       & {0.899 $\pm$ 0.064 (0.27 s/g)}  & 0.788 $\pm$  0.065 (0.23 s/g)         & {0.708 $\pm$  0.027 (1.58 s/g)}            \\
Erd\H{o}s' GNN (accurate)   &    0.915 $\pm$  0.060  (0.53 s/g)      & 0.799 $\pm$ 0.067 (0.46 s/g)          & 0.735 $\pm$  0.021 (6.68 s/g)               \\
RUN-CSP (fast)              & 0.833 $\pm$ 0.079 (0.27 s/g)           & 0.738 $\pm$ 0.067 (0.23 s/g)          & {0.771 $\pm$ 0.032 (1.84 s/g)}              \\
RUN-CSP (accurate)          & 0.892 $\pm$ 0.064 (0.51 s/g)           & 0.789 $\pm$ 0.053 (0.47 s/g)          & {0.804 $\pm$ 0.024 (5.46 s/g)}        \\ \midrule
Toenshoff-Greedy          &\textbf{ 0.924 $\pm$ 0.060 (0.02 s/g)}          & 0.816 $\pm$ 0.064 (0.02 s/g)          & \textbf{0.829 $\pm$ 0.027 (0.35 s/g)}
\\ \midrule
Gurobi 9.0 (0.1s)      &    0.889 $\pm$ 0.121 (0.18 s/g)                 & \textbf{0.795 $\pm$ 0.118 (0.16 s/g)} & 0.697 $\pm$ 0.033 (1.17 s/g) \\
Gurobi 9.0 (0.5s)      &  \textbf{0.962 $\pm$ 0.076 (0.34 s/g)}          & \textbf{0.855 $\pm$ 0.083 (0.31 s/g)} & 0.697 $\pm$ 0.033 (1.54 s/g) \\
Gurobi 9.0 (1.0s)      &  \textbf{0.980 $\pm$ 0.054 (0.45 s/g)}          & \textbf{0.872 $\pm$ 0.070 (0.40 s/g)} & 0.705 $\pm $ 0.039 (2.05 s/g)\\
Gurobi 9.0 (5.0s)      &  \textbf{0.998 $\pm$ 0.010 (0.76 s/g)}          & \textbf{0.884 $\pm$ 0.062 (0.68 s/g)} & 0.790 $\pm$ 0.285 (6.01 s/g)\\
Gurobi 9.0 (20.0s)     & \textbf{0.999 $\pm$ 0.003 (1.04 s/g)}           & \textbf{0.885 $\pm$ 0.063 (0.96 s/g)} & 0.807 $\pm$ 0.134 (21.24 s/g)\\ 
\bottomrule\\
\end{tabular} 
}
\caption{{Hard maximum clique instances (RB). We report the approximation ratio (bigger is better) in the training and test set, whereas the rightmost column focuses on a different distribution consisting of graphs of different sizes. Execution time is measured in sec. per graph (s/g). Pareto-optimal solutions are in bold.} \vspace{-0.4cm}} 
\label{table:max-clique-hard}
\end{table}

\subsection{Results: local graph partitioning}
% ============================================================

The results of all methods and datasets are presented in Table~\ref{table:partitioning}. 
To compare fairly with previous works, we evaluate partitioning quality based on the measure of local conductance, $\phi(S) = \cut{S}/\vol{S}$, even though our method only indirectly optimizes conductance. 
Nevertheless, Erd\H{o}s' GNN outperforms all previous algorithms by a considerable margin. We would like to stress that this result is not due to poor usage of previous methods: we rely on a well-known implementation~\citep{fountoulakis2018short} and select the parameters of all non-neural baselines by grid-search on a held-out validation set. We also do not report performance when a method (Pagerank-Nibble) returns the full graph as a solution~\cite{wang2017capacity}.  

It is also interesting to observe that, whereas all neural approaches perform well, GNN trained with a probabilistic loss attains better conductance across all datasets. We remind the reader that all three GNNs feature identical architectures and that the L1 and L2 loss functions are smooth relaxations that are heavily utilized in partitioning problems \cite{bresson2013multiclass}.
Furthermore, due to its high computational complexity and the extra overhead that is incurred when constructing the problem instances for large graphs, Gurobi performed poorly
in all but the smallest graphs.We argue that the superior solution quality of Erd\H{o}s' GNN serves as evidence for the benefit of our unsupervised framework. %

% We also note that Pagerank-Nibble can has been known to spread probability mass too aggressively and often returned the full graph as a solution; thus no results are listed for FACEBOOK . It should be noted that the performance of the local clustering methods often relies on th initilization seeds lying in a good conductance cluster of the graph. This explains the poor performance of SimpleLocal and MQI.

\begin{table}
\centering
\resizebox{0.88\textwidth}{!}{
\begin{tabular}{l c c c}     \toprule
                 &  SF-295                      &  FACEBOOK                      &  TWITTER \\ \midrule
Erd\H{o}s' GNN   & \textbf{0.124 $\pm$ 0.001 (0.22 s/g)} & \textbf{0.156 $\pm$ 0.026 (289.28 s/g)} &  \textbf{0.292 $\pm$ 0.009 (6.17 s/g)} \\ 
L1 GNN           & 0.188 $\pm$ 0.045 (0.02 s/g) & 0.571 $\pm$ 0.191 (13.83 s/g)  & \textbf{0.318 $\pm$ 0.077 (0.53 s/g)} \\ 
L2 GNN           & \textbf{0.149 $\pm$ 0.038 (0.02 s/g)} & \textbf{0.305 $\pm$ 0.082 (13.83 s/g)}  & 0.388 $\pm$ 0.074 (0.53 s/g) \\ \midrule
Pagerank-Nibble  & 0.375 $\pm$ 0.001 (1.48 s/g) & \text{N/A}                     & 0.603 $\pm$ 0.005 (20.62 s/g) \\
CRD              & 0.364 $\pm$ 0.001 (0.03 s/g) & 0.301 $\pm$ 0.097 (596.46 s/g) & 0.502 $\pm$ 0.020 (20.35 s/g) \\ 
MQI              & 0.659 $\pm$ 0.000 (0.03 s/g) & 0.935 $\pm$ 0.024 (408.52 s/g) & 0.887 $\pm$ 0.007 (0.71 s/g) \\
Simple-Local     & 0.650 $\pm$ 0.024 (0.05 s/g) & 0.955 $\pm$ 0.019 (404.67 s/g) & 0.895 $\pm$ 0.008 (0.84 s/g) \\ %\midrule
%Gurobi 9.0 (0.1s)    & 0.106 $\pm$ 0.000 (0.09 s/g) & 0.975 $\pm$ 0.000 (129.61 s/g) & 0.626 $\pm$ 0.010 (0.57 s/g) \\
%Gurobi 9.0 (1s)      & 0.105 $\pm$ 0.000 (0.09 s/g) & 0.975 $\pm$ 0.000 (129.42 s/g) & 0.576 $\pm$ 0.009 (1.82 s/g) \\
\midrule
Gurobi (10s)   & \textbf{0.105 $\pm$ 0.000 (0.16 s/g)}   & 0.961 $\pm$ 0.010 (1787.79 s/g)         & 0.535 $\pm$ 0.006 (52.98 s/g) \\ 
\bottomrule \\
\end{tabular} }
%\vspace{-0.2cm}
\caption{Cluster conductance on the test set (smaller is better) and execution time measured in sec. per graph. Pareto-optimal solutions are in bold.} %From left to right, all methods were rerun with 10, 100, and 25 random seeds per graph and only the best cluster was kept.   \vspace{-0.7cm}} 
\label{table:partitioning}
\end{table}

\section{Conclusion}
We have presented a mathematically principled framework for solving constrained combinatorial problems on graphs that utilizes a probabilistic argument to guarantee the quality of its solutions. %Our method is flexible as it allows us to incorporate constraints in different ways to the problem formulation and using our sequential decoding we can guarantee that the solutions obey said constraints. 
As future work, we would like to explore different avenues in which the sequential decoding could be accelerated. We aim to expand the ability of our framework to incorporate different types of constraints. Though we can currently support constraints where node order is not necessarily important (e.g., clique, cover, independent set), we would like to determine whether it is possible to handle more complex constraints, e.g., relating to trees or paths~\citep{ijcai2019-303}. %the conditional expectation can be costly to evaluate on large graphs. 
Overall, we believe that this work presents an important step towards solving CO problems in an unsupervised way and opens up the possibility of further utilizing techniques from combinatorics and the theory of algorithms in the field of deep learning.

\section{Broader impact}
% =================================================

This subfield of deep learning that our work belongs to is still in its nascent stages, compared to others like computer vision or translation. Therefore, we believe that it poses no immediate ethical or societal challenges. However, advances in combinatorial optimization through deep learning can have significant long term consequences. 
Combinatorial optimization tasks are important in manufacturing and transportation. The ability to automate these tasks will likely lead to significant improvements in productivity and efficiency in those sectors which will affect many aspects of everyday life. On the other hand, these tasks would be otherwise performed by humans which means that such progress may eventually lead to worker displacement in several industries. Combinatorial optimization may also lead to innovations in medicine and chemistry, which will be beneficial to society in most cases.

Our work follows the paradigm of unsupervised learning which means that it enjoys some advantages over its supervised counterparts. The lack of labeled instances implies a lack of label bias. Consequently, we believe that unsupervised learning has the potential to avoid many of the issues (fairness, neutrality) that one is faced with when dealing with labeled datasets. That does not eliminate all sources of bias in the learning pipeline, but it is nonetheless a step towards the right direction.

Finally, we acknowledge that combinatorial optimization has also been widely applied in military operations. However, even though this is not the intention of many researchers, we believe that it is just a natural consequence of the generality and universality of the problems in this field. Therefore, as with many technological innovations, we expect that the positives will outweigh the negatives as long as the research community maintains a critical outlook on the subject. Currently, the state of the field does not warrant any serious concerns and thus we remain cautiously optimistic about its impact in the world.
% ================================================
\section*{Acknowledgements}
We are grateful to the anonymous reviewers and Stefanie Jegelka for the helpful feedback and corrections. We would also like to thank the Swiss National Science Foundation for supporting this work in the context of the project \enquote{Deep Learning for Graph-Structured Data} (grant number PZ00P2 179981).
% ================================================
{
\small
\bibliography{bibliography}
}
% =================================================

% =================================================
\newpage
\appendix
% =================================================

\section{Visual demonstration}

\begin{figure}[h]
    \centering
    \includegraphics[width=1\textwidth]{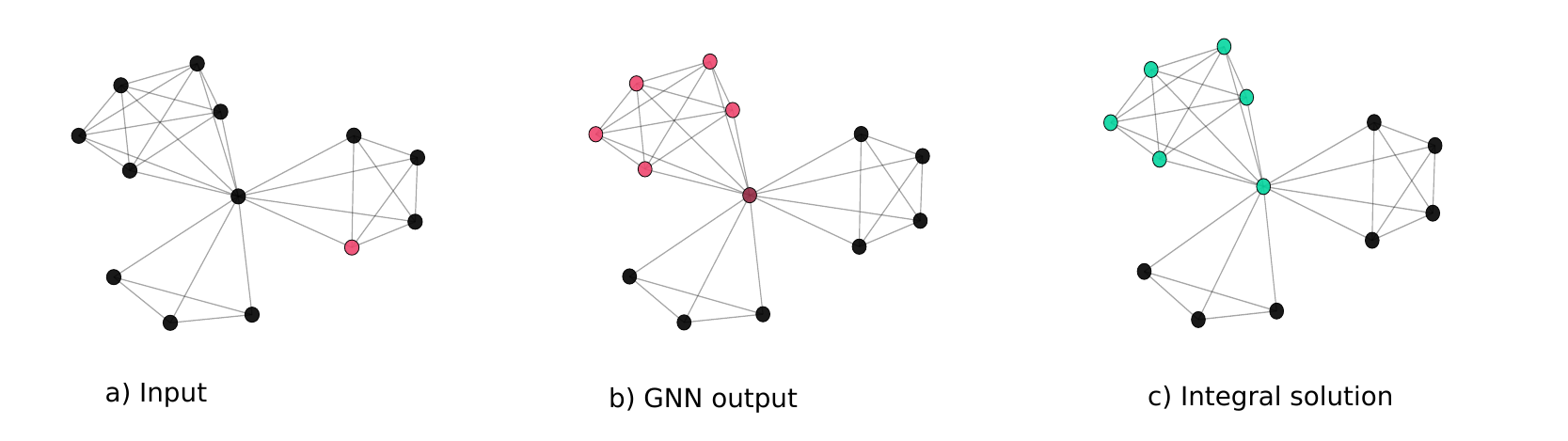}
    \caption{Illustration of our approach in a toy instance of the maximum clique problem from the IMDB dataset. a) A random node is selected to act as a `seed'. b) Erd\H{o}s' GNN outputs a probability distribution over the nodes (color intensity represents the probability magnitude) by exploring the graph in the vicinity of the seed. c) A set is sequentially decoded by starting from the node whose probability is the largest and iterating with the method of conditional expectation. The identified solution is guaranteed to obey the problem constraints, i.e., to be a clique.} 
    \label{fig:explanationl}
\end{figure}

Figure~\ref{fig:explanationl} provides a visual demonstration of the input and output of Erd\H{o}s' GNN in a simple instance of the maximum clique problem. 

We would like to make two observations. The first has to do with the role of the starting seed in the probability assignment produced by the network. In the maximum clique problem, we did not require the starting seed to be included in the solutions. This allowed the network to flexibly detect maximum cliques within its receptive field without being overly constrained by the random seed selection.  This is illustrated in the example provided in the figure, where the seed is located inside a smaller clique and yet the network is able to produce probabilities that focus on the largest clique. On the other hand, in the local graph partitioning problem we forced the seed to always lie in the identified solution---this was done to ensure a fair comparison with previous methods. Our second observation has to do with the sequential decoding process. It is encouraging to notice that, even though the central hub node has a considerably lower probability than the rest of the nodes in the maximum clique, the method of conditional expectation was able to reliably decode the full maximum clique.

% =================================================
\section{Experimental details}
\label{app:details}
% =================================================

% -------------------------------------
\subsection{Datasets}
% -------------------------------------

The following table presents key statistics of the datasets that were used in this study:

\begin{table}[h] 
\centering
\resizebox{1\textwidth}{!}{
\begin{tabular}{l cccccccc} \toprule
                  & IMDB  & COLLAB  & TWITTER  & RB (Train)&RB (Test)  & RB (Large Inst.) & SF-295 & FACEBOOK \\ \midrule
nodes              & 19.77 & 74.49   & 131.76   &  216.673 &  217.44        &        1013.25           & 26.06  & 7252.71   \\
edges              & 96.53 & 2457.78 & 1709.33  &   22852 & 22828 &   509988.2   & 28.08  & 276411.19   \\
reduction time     & 0.0003 &  0.006  & 0.024   & 0.018  & 0.018       &  0.252          & --    &  --     \\ 
number of test graphs & 200 & 1000 & 196 & 2000 & 500 & 40 & 8055 & 14 \\ \bottomrule \\
\end{tabular}} 
\caption{Average number of nodes and edges for the considered datasets. Reduction time corresponds to the average number of seconds needed to reduce a maximum clique instance to a maximum independent instance. Number of test graphs refers to the number of graphs that the methods were evaluated on, in a given dataset.} \label{table:datasetstats}
\end{table} 

To speed up computation and training, for the Facebook dataset, we kept graphs consisting of at most 15000 nodes (i.e., 70 out of the total 100 available graphs of the dataset). %We did this to speed up computation and training. % and also helped us avoid the large memory costs incurred when training with those graphs.

The RB test set can be downloaded from the following link: \url{https://www.dropbox.com/s/9bdq1y69dw1q77q/cliques_test_set_solved.p?dl=0}.
The latter was generated using the procedure described by \citet{xu2007benchmarks}. We used a python implementation by \citet{toenshoff2019run} that is available on the RUN-CSP repository: \url{https://github.com/RUNCSP/RUN-CSP/blob/master/generate_xu_instances.py}.
Since the parameters of the original training set were not available, we selected a set of initial parameters such that the generated dataset resembles the original training set.
As seen in Table~\ref{table:datasetstats}, the properties of the generated test set are close to those of the training set. Specifically, the training set contained graphs whose size varied between 50 and 500 nodes and featured cliques of size 5 to 25. The test set was made out of graphs whose size was between 50 and 475 nodes and contained cliques of size 10 to 25. These minor differences provide a possible explanation for the drop in test performance of all methods (larger cliques tend to be harder to find).

All other datasets are publicly available. 

% An important detail is that we restricted the size of the maximum cliques to lie within the [10,25] interval, compared to the [5,25] in the training set. The graphs in the test set have sizes in the [50,475] interval, while for the training set the graphs have sizes in the [50,500] interval. These slight modifications helped us avoid duplicate instances between training and test sets and gave emphasis on larger cliques which tend to be harder. These differences partly explain the drop in performance of all methods, compared to the training set where the maximum clique can be smaller.

% -------------------------------------------------
\subsection{Neural network architecture}
% -------------------------------------------------

In both problems, Erd\H{o}s' GNN and our own neural baselines were given as node features a one-hot encoding of a random node from the input graph. % 
For the local graph partitioning setting, our networks consisted of 6 GIN layers followed by a multi-head GAT layer. The depth was kept constant across all datasets. We employed skip connections and batch-normalization at every layer. 
For the maximum clique problem, we also incorporated graph size normalization for each convolution, as we found that it improved optimization stability. The networks in this setting did not use a GAT layer, as we found that multi-head GAT had a negative impact on the speed/memory of the network, while providing only negligible benefits in accuracy.
Furthermore, locality was enforced after each layer by masking the receptive field. That is, after 1 layer of convolution only 1-hop neighbors were allowed to have nonzero values, after 2 layers only 2-hop neighbors could have nonzero values, etc. 
The output of the final GNN layer was passed through a two layer perceptron giving as output one value per node. The aforementioned numbers were re-scaled to lie in $[0,1]$ (using a graph-wide min-max normalization) and were interpreted as probabilities $p_1, \ldots, p_n$. In the case of local graph partitioning, the forward-pass was concluded by the appropriate re-scaling of the probabilities (as described in Section~\ref{subsubsec:box_constraint}).  %As another measure that enforces locality, we set the probability of the input starting seed node to 1 for Erd\H{o}s GNN and the neural baselines. This guarantees that the starting node will be included in the solution.

% -------------------------------------
\subsection{Local graph partitioning setup}
\label{app:details_lgp}
% -------------------------------------

Following the convention of local graph clustering algorithms, for each graph in the test set we randomly selected $d$ nodes of the input graph to act as cluster \textit{seeds}, where $d=10,30$, and $100$ for SF-295, TWITTER, and FACEBOOK, respectively. Each method was run once for each seed resulting in $d$ sets per graph. We obtained one number per seed by averaging the conductances of the graphs. Table~\ref{table:partitioning} reports the mean and standard deviation of these numbers.
%\footnote{Please note that the caption of Table~\ref{table:partitioning} incorrectly reports that $d = 25$ seeds were used for the TWITTER dataset and that the best conductance was kept for each graph. %The correct number of seeds that was used is $d=30$. 
The correct procedure is the one described here.%}

%to complete, explain locality, differences from Gurobi%
The volume-constrained graph partitioning formulation can be used to minimize conductance as follows: Perform grid search over the range of feasible volumes and create a small interval around each target volume. Then, solve a volume-constrained partitioning problem for each interval, and return the set of smallest conductance identified.   

We used a fast and randomized variant of the above procedure with all neural approaches and Gurobi (see Section~\ref{subsec:localgraphpartitioning} for more details). Specifically, for each seed node we generated a random volume interval within the receptive field of the network, and solved the corresponding constrained partitioning problem. Our construction ensured that the returned sets always contained the seed node and had a controlled volume. For L1 and L2 GNN, we obtained the set by sampling from the output distribution. We drew 10 samples and kept the best. We found that in contrast to flat thresholding (like in the maximum clique), sampling yielded better results in this case. 

For the parameter search of local graph clustering methods, we found the best performing parameters on a validation set via grid search when that was appropriate. For CRD, we searched for all the integer values in the [1,20] interval for all 3 of the main parameters of the algorithm. For Simple Local, we searched in the [0,1] interval for the locality parameter. Finally, for Pagerank-Nibble we set a lower bound on the volume that is 10 \% of the total graph volume. It should be noted, that while local graph clustering methods achieved inferior conductance results, they do not require explicit specification of a receptive field which renders them more flexible.

% Our neural network is trained with randomly generated target volumes during training. The target volumes lie in the $[0.1*, ]$

\subsection{Hardware and software}

All methods were run on an Intel Xeon Silver 4114 CPU, with 192GB of available RAM. The neural networks were executed on a single RTX TITAN 25GB graphics card. The code was executed on version 1.1.0 of PyTorch and version 1.2.0 of PyTorch Geometric.
% % -------------------------------------
% \subsection{Pre-trained Models}
% % -------------------------------------

% Pre-trained models of Erd\H{o}s' GNN for the maximum clique and the constrained minimum cut respectively can be downloaded from the following links:\\
% \url{https://www.dropbox.com/sh/mdsjrcg9gch8dti/AADW0UUcQMUkChz8SZNXYGnVa?dl=0} \\
%      \url{https://www.dropbox.com/sh/z00ictftyxx3ipf/AADirtiMIwI3_sxCep5GzJf_a?dl=0}
% ===============================================================
\section{Additional results}
% ===============================================================

% -------------------------------------
\subsection{Maximum clique problem}
% -------------------------------------

The following experiments provide evidence that both the learning and decoding phases of our framework are important in obtaining valid cliques of large size.   

\subsubsection{Constraint violation}

Table~\ref{table:max-clique-violation} reports the percentage of instances in which the clique constraint was violated in our experiments. Neural baselines optimized according to penalized continuous relaxations struggle to detect cliques in the COLLAB and TWITTER datasets, whereas Erd\H{o}s' GNN always respected the constraint. 

\begin{table}[h!]
\centering
\resizebox{0.65\textwidth}{!}{
\begin{tabular}{l c c c c}    \toprule
\emph{}                    & {IMDB}        & COLLAB       & TWITTER      & RB (all datasets) \\\midrule
Erd\H{o}s' GNN (fast)      & \textbf{0\%}  & \textbf{0\%} & \textbf{0\%} & \textbf{0\%} \\ 
Erd\H{o}s' GNN (accurate)  & \textbf{0\%}  & \textbf{0\%} & \textbf{0\%} & \textbf{0\%} \\ 
Bomze GNN                  & \textbf{0\%}  & 11.8\%       & 78.1\%       & -- \\
MS GNN                     & 1\%           & 15.1\%       & 84.7\%       & -- \\
\bottomrule \\
\end{tabular} }
\caption{Percentage of test instances where the clique constraint was violated.\vspace{-.4cm}}
\label{table:max-clique-violation}
\end{table}

Thus, decoding solutions by the method of conditional expectation is crucial to ensure that the clique constraint is always satisfied. 

\subsubsection{Importance of learning}

We also tested the efficacy of the learned probability distributions produced by our GNN on the Twitter dataset. We sampled multiple random seeds and produced the corresponding probability assignments by feeding the inputs to the GNN. These were then decoded with the method of conditional expectation and the best solution was kept. To measure the contribution of the GNN, we compared to random uniform probability assignments on the nodes. In that case, instead of multiple random seeds, we had the same number of multiple random uniform probability assignments. Again, these were decoded with the method of conditional expectation and the best solution was kept. The results of the experiment can be found in Table \ref{table:ablation}.
\begin{table}[h!]
\centering
\resizebox{0.45\textwidth}{!}{
\begin{tabular}{l c c c} \toprule 
 & Erd\H{o}s' GNN        & U $\sim$ [0,1]            &  \\ \midrule
1  sample     & 0.821 $\pm$ 0.222 & 0.513 $\pm$   0.266 &  \\
3  samples       & 0.875 $\pm$ 0.170 & 0.694 $\pm$ 0.210    &  \\
5  samples       & 0.905 $\pm$ 0.139 & 0.760  $\pm$  0.172  &  \\ \bottomrule \\ 
\end{tabular}}
\caption{Approximation ratios with sequential decoding using the method of conditional expectation on the twitter dataset. The second column represents decoding with the probabilities produced by the GNN. The third column shows the results achieved by decoding random uniform probability assignments on the nodes. \vspace{-0.1cm}}
\label{table:ablation}
\end{table}  

As observed, the cliques identified by the trained GNN were significantly larger than those obtained when decoding a clique from a random probability assignment.

% -------------------------------------
\subsection{Local graph partitioning}
\label{subsec:localgraphpartitioning}
% -------------------------------------
We also attempted to find sets of small conductance using Gurobi. To ensure a fair comparison, we mimicked the setting of Erd\H{o}s' GNN and re-run the solver with three different time-budgets, making sure that the largest budget exceeded our method's running time by approximately one order of magnitude. %
We used the following integer-programming formulation of the constrained graph partitioning problem: 
\begin{align}
    \min_{x_1, \ldots, x_n \in \{0,1\} } &\sum_{(v_{i},v_{j}) \in E} (x_{i} -x_{j})^{2}\\
    \text{subject to} & \quad   \left(1-\frac{1}{4}\right) \, \text{vol} \leq \sum_{v_{i}\in V} x_{i}d_{i}  \leq \left(1+\frac{1}{4}\right) \, \text{vol} \quad \text{and} \quad x_{s} = 1. \nonumber
\end{align}
Above, $\text{vol}$ is a target volume and $s$ is the index of the seed node (see explanation in Section~\ref{app:details_lgp}). Each binary variable $x_i$ is used to indicate membership in the solution set. In order to encourage local solutions on a global solver like Gurobi, the generated target volumes were set to lie in an interval that is attainable within a fixed receptive field (identically to the neural baselines). Additionally, the seed node $v_{s}$ was also required to be included in the solution. The above choices are consistent with the neural baselines and the local graph partitioning setting.

The results are shown in Table~\ref{table:partitioning_gurobi}. Due to its high computational complexity, Gurobi performed poorly in all but the smallest instances. In the FACEBOOK dataset, which contains graphs of 7k nodes on average, Erd\H{o}s' GNN was impressively able to find sets of more than 6$\times$ smaller conductance, while also being 6$\times$ faster.

\begin{table}[h!]
\centering
\resizebox{0.88\textwidth}{!}{
\begin{tabular}{l c c c}     \toprule
                 &  SF-295                      &  FACEBOOK                      &  TWITTER \\ \midrule
Gurobi (0.1s)  & {0.107 $\pm$  0.000 (0.16 s/g)}         & 0.972 $\pm$ 0.000 (799.508 s/g)         & 0.617 $\pm$ 0.012 (3.88 s/g) \\ 
Gurobi (1s)    & {0.106 $\pm$  0.000 (0.16 s/g)}         & 0.972 $\pm$ 0.000 (893.907 s/g)         & 0.544 $\pm$ 0.007 (12.41 s/g)\\ 
Gurobi (10s)   & \textbf{0.105 $\pm$ 0.000 (0.16 s/g)}   & 0.961 $\pm$ 0.010 (1787.79 s/g)         & 0.535 $\pm$ 0.006 (52.98 s/g) \\ \midrule
Erd\H{o}s' GNN & {0.124 $\pm$ 0.001 (0.22 s/g)}          & \textbf{0.156 $\pm$ 0.026 (289.28 s/g)} & \textbf{0.292 $\pm$ 0.009 (6.17 s/g)} \\ 
%Gurobi 9.0 (0.1s)    & 0.106 $\pm$ 0.000 (0.09 s/g) & 0.975 $\pm$ 0.000 (129.61 s/g) & 0.626 $\pm$ 0.010 (0.57 s/g) \\
%Gurobi 9.0 (1s)      & 0.105 $\pm$ 0.000 (0.09 s/g) & 0.975 $\pm$ 0.000 (129.42 s/g) & 0.576 $\pm$ 0.009 (1.82 s/g) \\
\bottomrule \\
\end{tabular} }
%\vspace{-0.2cm}
\caption{Average conductance of sets identified by Gurobi and Erd\H{o}s' GNN (these results are supplementary to those of Table~\ref{table:partitioning}).}
\label{table:partitioning_gurobi}
\end{table}

It should be noted that the time budget allowed for Gurobi only pertains to the \emph{optimization time} spent (for every seed). There are additional costs in constructing the problem instances and their constraints for each graph. These costs become particularly pronounced in larger graphs, where setting up the problem instance takes more time than the allocated optimization budget. We report the total time cost in seconds per graph (s/g).

\section{Deferred technical arguments}
% =================================================

% =================================================
\subsection{Proof of Theorem~\ref{theorem:probabilistic_penalty_certificate}}
% =================================================

In the constrained case, the focus is on the probability $P(\{f_{}(S;G) < \epsilon\} \cap \{S \in \Omega\})$.
Define the following probabilistic penalty function:
\begin{align}
    f_{p}(S;G) = f(S;G) + \cvec{1}_{S \notin \Omega} \, \beta,  \label{eq:probpen}
\end{align} 
where $\beta$ is any number larger than $\max_S \{f(S;G)\}$.
The key observation is that, if $\ell(\mathcal{D}, G) = \epsilon < \beta$, then there must exist a valid solution of cost $\epsilon$. It is a consequence of $f(S;G) > 0$ and $\beta$ being an upper bound of $f$ that 
\begin{align}
    P(f_{p}(S;G) < \epsilon ) = P(f_{}(S;G) < \epsilon \cap S \in \Omega).
\end{align}
Similar to the unconstrained case, for a non-negative $f$, Markov's inequality can be utilized to bound this probability:
\begin{align}
     P( \{f_{}(S;G) < \epsilon\} \cap \{S \in \Omega\})
     &= P(f_{p}(S;G) < \epsilon)  \notag \\
     &> 1-\frac{1}{\epsilon} \E{f_{p}(S;G)} \notag \\
     &= 1- \frac{1}{\epsilon} \left( \E{f_{}(S;G)} + \E{\cvec{1}_{S \notin \Omega}\, \beta} \right)\notag \\
    %  &= 1- \frac{1}{\epsilon} \left( \E{f_{o}(S;G)} + P(S \in \Omega)\E{\cvec{1}_{S \notin \Omega} \max(f_{c}(S;G))|S \in \Omega} + P(S \notin \Omega) \E{ \cvec{1}_{S \notin \Omega} \max(f_{c}(S;G))|S \notin \Omega} \right)\notag \\     
     &= 1- \frac{1}{\epsilon} \left( \E{f_{}(S;G)} + P(S \notin \Omega) \, \beta \right).
\end{align}
The theorem claim follows from the final inequality. 

% ======================================================================
\subsection{Iterative scheme for non-linear re-scaling}
\label{app:iterative_scheme}
% ======================================================================

Denote by $\mathcal{D}^{0}$ the distribution of sets predicted by the neural network and let $p_1^{0}, \ldots, p_n^{0}$ be the probabilities that parameterize it. 
We aim to re-scale these probabilities such that the constraint is satisfied in expectation: 
$$
     \sum_{v_i \in V} a_i p_i  = \frac{b_l + b_h}{2}, \quad \text{where} \quad p_i = \clamp{c \, p_i^{0}}{0}{1}   \quad \text{and} \quad  c \in \mathbb{R}.
$$
This can be achieved by iteratively applying the following recursion: 
\begin{align*}
    p_i^{\tau+1} \leftarrow \clamp{ c^\tau p_i^{\tau}}{0}{1}, \quad \text{with} \quad c^\tau = \frac{b - \sum_{v_i \in Q^{\tau}} a_i }{ \sum_{v_i \in V \setminus Q^\tau} a_i p_i^{\tau}} \quad \text{and} \quad Q^\tau = \{ v_i \in V \, : \, p_i^\tau = 1\},
\end{align*}
where $b = \frac{b_l + b_h}{2}$.

The fact that convergence occurs can be easily deduced. Specifically, consider any iteration $\tau$ and let $Q^\tau$ be as above. If $p_i^{\tau+1} < 1$ for all ${v_i \in V \setminus Q^\tau}$, then the iteration has converged. Otherwise, we will have $Q^\tau \subset Q^{\tau+1}$. From the latter, it follows that in every $\tau$ (but the last), set $Q^\tau$ must expand until either $\clamp{ c^\tau p_i^{\tau}}{0}{1} = b$ or $Q^\tau = V$. The latter scenario will occur if $\sum_{v_i \in V} a_i \leq b$.          

% ======================================================================
\subsection{Proof of Theorem~\ref{theorem:linear-box-constraint}}
% ======================================================================

Set $b = (b_l + b_h)/2$ and $\delta = (b_h - b_l)/2$. By Hoeffding's inequality, the probability that a sample of $\mathcal{D}$ will lie in the correct interval is:
$$
    \Prob{ \left|\sum_{v_i \in S} a_i - \E{\sum_{v_i \in S} a_i} \right| \leq \delta } 
    = \Prob{\left|\sum_{v_i \in S} a_i - b \right| \leq \delta } 
    \geq 1- 2 \exp{\left(-\frac{2\delta^2 }{ \sum_{i} a_i^2 } \right)}.  
$$
We can combine this guarantee with the unconstrained guarantee by taking a union bound over the two events:
\begin{align*}
    \Prob{ f(S;G) < \ell(\mathcal{D}, G) \text{ AND } \sum_{v_i \in S} a_i \in [b_l , b_h]} \\
    &\hspace{-5cm}= 1 - \Prob{ f(S;G) \geq \ell(\mathcal{D}, G) \text{ OR } \sum_{v_i \in S} a_i \notin [b_l , b_h] } \\
    &\hspace{-5cm}\geq 1 - \Prob{ f(S;G) \geq \ell(\mathcal{D}, G)} - \Prob{\sum_{v_i \in S} a_i \notin [b_l , b_h] } \\    
    &\hspace{-5cm}\geq t - 2 \exp{\left(-\frac{2\delta^2 }{ \sum_{i} a_i^2 } \right)} 
\end{align*}
The previous is positive whenever $t > 2 \exp{\left(- 2\delta^2 /  (\sum_{i} a_i^2 ) \right)}$. 

% ============================================================
\subsubsection{Proof of Corollary~\ref{corrolary:max_clique}}
% ============================================================

To ensure that the loss function is non-negative, we will work with the translated objective function 
$f(S;G) = \gamma - w(S)$, where the term $\gamma$ is any upper bound of $ w(S) $ for all $S$. 

Theorem~\ref{theorem:probabilistic_penalty_certificate} guarantees that if 
\begin{align}
     \E{f(S;G)} + P(S \notin \Omega) \, \beta \leq \ell_{\text{clique}}(\mathcal{D}, G) \leq \epsilon(1-t)
\end{align}
and as long as $\max_S f(S;G) = \gamma - \min_S w(S) \leq \gamma \leq \beta$, then with probability at least $t$, set $S^* \sim \mathcal{D}$ satisfies $\gamma- \epsilon <w(S^*)$. 

Denote by $x_i$ a Bernoulli random variable with probability $p_i$.
It is not difficult to see that 
\begin{align}
    \E{w(S)} 
    &= \E{\sum_{(v_i,v_j) \in E} w_{ij} x_{i} x_{j}}  = \sum_{(v_i,v_j) \in E} w_{ij} p_{i} p_{j}  \label{eq:expected_weight}
\end{align}
% 
% $
%     \E{f(S;G)} = \gamma - \E{w(S)},  
% $    
% with

We proceed to bound $P(S \notin \Omega_{\text{clique}})$. Without loss of generality, suppose that the edge weights have been normalized to lie in $[0,1]$. We define $\bar{w}(S)$ to be the weight of $S$ on the complement graph: 
$$
    \bar{w}(S) \triangleq \sum_{v_i,v_j \in S} \mathbf{1}_{ \{(v_i,v_j) \notin E \} }  
$$
By definition, we have that 
$
    \Prob{ S \notin \Omega_{\text{clique}}} = \Prob{\bar{w}(S) \geq 1}.
$ 
Markov's inequality then yields
\begin{align}
    \Prob{ S \notin \Omega_{\text{clique}}} 
    \leq \E{\bar{w}(S)} \notag 
    &= \E{ \frac{|S|(|S|-1)}{2} } - \E{w(S)} \notag \\
    &\hspace{-3cm}= \frac{1}{2} \E{\left(\sum_{v_i \in V} x_i\right)^2 - \sum_{v_i \in V} x_i}  - \E{w(S)} \notag \\
    &\hspace{-3cm}= \frac{1}{2}\sum_{v_i\neq v_j } \E{x_i x_j} + \frac{1}{2} \sum_{v_i \in V} \E{x_i^{2}} - \sum_{v_i \in V} \E{x_i} - \frac{1}{2} \E{w(S)} \notag\\
    &\hspace{-3cm}= \frac{1}{2} \sum_{v_i\neq v_j } p_i p_j +  \frac{1}{2} \sum_{v_i \in V} p_i - \frac{1}{2} \sum_{v_i \in V} p_i - \E{w(S)} 
    = \frac{1}{2} \sum_{v_i\neq v_j } p_i p_j - \E{w(S)}.
\end{align}
It follows from the above derivations that 
\begin{align}
      \gamma - \E{w(S)} + P(S \notin \Omega) \, \beta
      &\leq \gamma - \E{w(S)} + \frac{\beta}{2} \sum_{v_i\neq v_j } p_i p_j - \beta\E{w(S)} \notag \\
      &= \gamma - (1+\beta) \E{w(S)} +\frac{\beta}{2} \sum_{v_i\neq v_j } p_i p_j \notag \\
      &= \gamma - (1+\beta) \sum_{(v_i,v_j) \in E} w_{ij} p_{i} p_{j} +\frac{\beta}{2} \sum_{v_i\neq v_j } p_i p_j.
\end{align}
The final expression is exactly the probabilistic loss function for the maximum clique problem. 

\subsection{Proof of Corollary~\ref{cor:min_cut}}

Denote by $S$  the set of nodes belonging to the cut, defined as
% We can express the absolute value of their pairwise differences in the following way:
%
$S = \{v_i \in V, \ \text{such that} \  x_i = 1 \}.$
Our first step is to re-scale the probabilities such that, in expectation, the following is satisfied: 
$$ 
    \E{\vol{S}} = \frac{v_l + v_h}{2}.
$$
This can be achieved by noting that the expected volume is 
\begin{align*}
    \E{\vol{S}} &= \E{\sum_{v_i \in V} d_i x_i} = \sum_{v_i \in V} d_i p_i 
\end{align*}
and then using the procedure described in Section~\ref{app:iterative_scheme}.

With the probabilities $p_1, \ldots, p_n$ re-scaled, we proceed to derive the probabilistic loss function corresponding to the min cut.

The cut of a set $S \sim \mathcal{D}$ can be expressed as 
\begin{equation}
\cut{S} 
  = \sum_{v_i \in S} \sum_{v_j \notin S} w_{ij}
  = \sum_{(v_i,v_j) \in E} w_{ij} z_{ij}, 
%   &= \sum_{i,j \in E} w_{ij} |x_{i} - x_{j}|    
\end{equation}
where $z_{ij}$ is a Bernoulli random variable with probability $p_i$ which is equal to one if exactly one of the nodes $v_i,v_j$ lies within set $S$. Formally,
\begin{align}
    z_{ij} = |x_{i}-x_{j}|  = \begin{cases}
                  1 \quad \text{with probability }   p_{i} -2 p_{i}p_{j}  + p_{j}  \\
                  0 \quad \text{with probability }   2p_{i}p_{j} - (p_{i}+p_{j})  + 1
     \end{cases}
\end{align}
It follows that the expected cut is given by
\begin{align*}
\E{\cut{S}}
  &= \sum_{(v_{i},v_{j}) \in E} w_{ij} \, \E{z_{ij}} \\  
  &= \sum_{(v_{i},v_{j}) \in E}w_{ij}(p_{i} - 2p_{i}p_{j}  + p_{j}) \\  
  &= \sum_{(v_{i},v_{j}) \in E} w_{ij}(p_{i} + p_{j}) - 2\sum_{(v_i,v_j) \in E} p_{i}p_{j} w_{ij}
  = \sum_{v_i \in V} d_i p_i - 2\sum_{(v_i,v_j) \in E} p_{i}p_{j} w_{ij}.
\end{align*}
We define, accordingly, the min-cut probabilistic loss as 
$$
    \ell_{\text{cut}}(\mathcal{D}; G) =  \sum_{v_i \in V} d_i p_i - 2\sum_{(v_i,v_j) \in E} p_{i}p_{j} w_{ij}
$$

Then, for any $t \in (0,1]$, Markov's inequality yields: 
$$
    \Prob{ \cut{S} < \frac{\ell_{\text{cut}}(\mathcal{D}; G)}{1-t}} > t
$$
The proof then concludes by invoking Theorem~\ref{theorem:linear-box-constraint}.

\end{document}